%% file: main.tex
\documentclass[sigconf,nonacm]{acmart}
\makeatletter
\@ACM@balancefalse
\makeatother

\let\cite\citep
\usepackage{xspace}
\usepackage{array}
\usepackage{multirow}
\usepackage{algorithm}
\usepackage{algorithmic}
\usepackage{tabularx}
\usepackage{listings}
\usepackage[most]{tcolorbox}
\usepackage{tikz}
\usetikzlibrary{arrows.meta,positioning,calc}
\hypersetup{hidelinks}
\newcommand{\paperfigure}[2][]{%
  \IfFileExists{#2}{%
    \includegraphics[#1]{#2}%
  }{%
    \fbox{%
      \begin{minipage}[c][0.16\textheight][c]{0.92\linewidth}
        \centering\small
        Figure asset not found:\par
        \path{#2}
      \end{minipage}%
    }%
  }%
}

\newcommand{\sys}{\mbox{\normalfont\textsc{SymTrace}}\xspace}
\newcommand{\ds}{\mbox{\normalfont\textsc{SymFail}}\xspace}
\newcounter{finding}
\newcommand{\finding}[1]{%
  \refstepcounter{finding}%
  \par\noindent\textbf{Finding \thefinding.} \textit{#1}%
  \par
}

\title{Repair or Resample? Rethinking Failure Debugging in LLM Multi-Agent Systems}
\author{Zhongwen Luan}
\authornote{These authors contributed equally.}
\affiliation{%
  \institution{East China Normal University}
  \city{Shanghai}
  \country{China}
}
\email{10245101408@stu.ecnu.edu.cn}

\author{Xiaoyu Zhang}
\authornotemark[1]
\affiliation{%
  \institution{Nanyang Technological University}
  \city{Singapore}
  \country{Singapore}
}

\author{Yue Yang}
\affiliation{%
  \institution{Xi'an University of Architecture and Technology}
  \city{Xi'an}
  \country{China}
}

\author{Jiongchi Yu}
\affiliation{%
  \institution{Nanyang Technological University}
  \city{Singapore}
  \country{Singapore}
}

\author{Xiaohong Chen}
\affiliation{%
  \institution{East China Normal University}
  \city{Shanghai}
  \country{China}
}

\author{Ming Hu}
\authornote{Corresponding author.}
\affiliation{%
  \institution{East China Normal University}
  \city{Shanghai}
  \country{China}
}

\renewcommand{\shortauthors}{Luan et al.}
\begin{document}
\begin{abstract}
As large language model (LLM)-based multi-agent systems (MASs) are increasingly applied to long-horizon complex tasks, their reliability has emerged as the core bottleneck hindering their real-world deployment.
Existing MAS debugging and repair methods typically rely on rerunning and resampling the entire execution trajectory.
However, a fundamental question remains to be answered: \textit{do these methods causally repair MAS failures or merely stochastically repair by leveraging the randomness of LLM sampling?}
To evaluate the effectiveness of MAS repair methods, we introduce \sys, a controlled evaluation framework that records the MAS execution trajectory and establishes intervention anchors.
During replay, it effectively reconstructs the execution before the anchor using recorded logs and only regenerates the downstream trajectory, thereby enabling the reliable reproduction of MAS failures.
We further construct the dataset \ds, comprising 536 human-annotated failure trajectories with graph-linked locations, categories, and trace evidence. 
Based on these foundations, we conduct a large-scale empirical study across three mainstream MAS frameworks.
Our findings reveal that existing unguided rerun methods are highly unreliable, exhibiting low failure reproduction and repair rates (only 67.97\% and 6.90\%, respectively).
Building upon these findings, we further explore the effectiveness of a symptom-driven intervention method, which successfully repairs 20.15\% of the failed cases (a 191.89\% improvement to state-of-the-art repair methods). 
This study aims to provide actionable insights for MAS debugging and repair research, paving the way for the robust deployment of multi-agent systems.
\end{abstract}

\maketitle

\noindent\textbf{Source code:}
\url{https://anonymous.4open.science/r/SymTrace-7234}
\par

\section{Introduction}

Large language model (LLM)-based multi-agent systems (MASs) coordinate specialized agents through message passing, shared context, and structured workflows~\cite{wu2023autogen,hong2024metagpt,qian2024chatdev,fourney2024magentic}. By distributing planning, information gathering, tool use, execution, and verification, MASs offer a promising foundation for long-horizon applications such as web interaction and software development~\cite{zhou2024webarena,yoran2024assistantbench,yang2024swe}. As these systems are deployed in increasingly complex and consequential settings, systematically understanding, reproducing, and repairing their failures becomes essential to reliability and safety.

\begin{figure}[t]
\centering
\paperfigure[
    width=\columnwidth,
    keepaspectratio
]{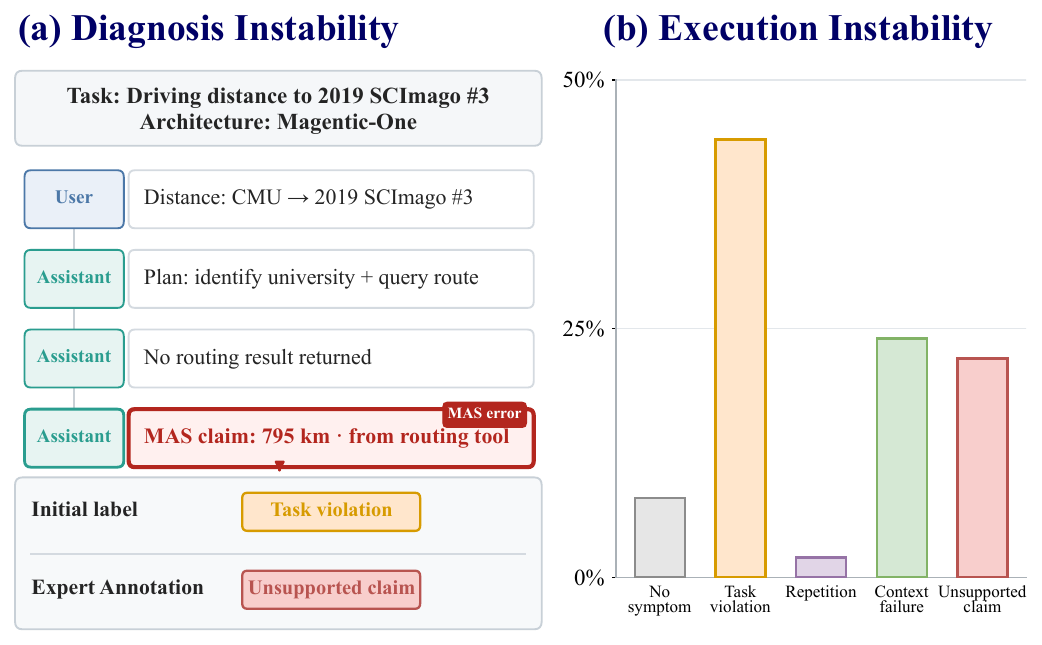}
\caption{Diagnosis and execution instability for the same MAS task.
{\scriptsize
(a) In a Magentic-One distance-query failure, the system makes a routing-derived claim despite receiving no routing result, leading to disagreement between the initial and expert diagnoses.
(b) Repeated stochastic reruns of the same task produce different failure types, while some exhibit no detected symptom.}}
\label{fig:intro}
\Description{A comparison of diagnosis disagreement and stochastic rerun outcomes for the same multi-agent-system task.}
\end{figure}

Existing research has advanced MAS reliability through benchmarks, failure diagnosis, and execution repair. Benchmarking and diagnostic studies collect execution trajectories, construct failure taxonomies, and localize responsible agents or steps~\cite{yao2024tau,ye2026claw,cemri2025multi,shah2026characterizing,deshpande2025trail,zhang2025agent}, while repair methods use rerunning, reflection, critic feedback, or trajectory-level guidance to generate new executions~\cite{madaan2023self,shinn2023reflexion,du2023improving,gou2024critic,zhao2026debugging,nanda2026wink,zhao2026agenttether}. However, two limitations remain. First, complete reruns resample upstream model decisions instead of holding the failure-producing execution fixed, making terminal success difficult to attribute to the applied repair. Second, task-level verdicts and automatically assigned categories may not identify the trace-localized behavior that actually requires intervention. Figure~\ref{fig:intro} illustrates how both the diagnosis and observed outcome can change without establishing that the recorded failure was corrected. A central question therefore remains: \textit{Do these methods causally repair MAS failures, or merely achieve stochastic recovery through LLM resampling?}

To fill this gap, we introduce \sys, a replay-oriented logging system, together with \ds, a human-annotated failure-trajectory dataset. During snapshot recording, \sys captures model and tool boundary interactions, organizes them into an event-dependency graph and observed execution order, and materializes the resulting records as a replay bundle. During replay, it restarts the native MAS, strictly matches each intercepted request, and injects the corresponding recorded result until reaching the designated intervention boundary. Live execution then resumes with the repair applied. This procedure holds the observable pre-intervention history fixed so that downstream changes can be attributed to the intervention rather than upstream resampling. Complementing this execution control, \ds is constructed from 200 WebArena-Verified Hard and AssistantBench tasks~\cite{zhou2024webarena,yoran2024assistantbench} executed with AG2~\cite{wu2023autogen}, CrewAI~\cite{crewai2026}, and Magentic-One~\cite{fourney2024magentic}. Of the resulting 600 trajectories, 536 evaluator-confirmed failures are retained and human-annotated with graph-linked failure nodes, symptom categories, supporting evidence, and final evidence-based annotations.

Using these two resources, we evaluate task-level repair and node-level controls across 536 failures. \sys increases per-execution failure reproduction from 67.97\% to 80.78\% and consistent reproduction across three executions from 41.42\% to 52.43\%. Task-level regeneration repairs at most 6.90\% of failures within three attempts, whereas \textit{Suspicious-Node Intervention} repairs 20.15\% with a single intervention (2.92$\times$ the rate of the strongest task-level baseline) and performs best across all three MASs. These results demonstrate the value of controlled execution and localized evidence over repeated stochastic regeneration.

Our contributions include:
\ding{182} We design and implement \sys, a logging system that records each MAS execution as an event-dependency snapshot and supports verifiable prefix reconstruction and replay, enabling intervention without resampling the preceding execution.
\ding{183} We construct \ds, a human-annotated dataset including 536 human-annotated failure trajectories from WebArena-Verified Hard and AssistantBench, which forms the foundation of the MAS debugging study.
\ding{184} We conduct a large-scale study across three MASs measuring how reliably existing methods reproduce and repair recorded failures.
Guided by our findings, we propose a symptom-driven intervention method that localizes replayable anchors and applies evidence-conditioned repair, improving over the strongest task-level baseline by 191.89\%.
\ding{185} We release \sys, \ds, and our experimental results to support reproducibility and future MAS debugging and repair research.


\section{Background \& Related Work}

\subsection{LLM-Based Multi-Agent Systems}

LLM-based multi-agent systems (MASs) coordinate multiple LLM-powered agents to accomplish a shared task. Each agent is typically assigned a role, instructions, context, and a set of tools, while a concrete execution proceeds through model calls, inter-agent messages, intermediate artifacts, and interactions with external environments. The coordination architecture determines how tasks are decomposed, which agent acts next, and how intermediate state is transferred. The systems evaluated in our study cover three representative designs. AG2 supports programmable conversational interactions among customizable agents~\cite{wu2023autogen}; CrewAI organizes role-specialized agents through sequential or hierarchical task processes~\cite{crewai2026}; and Magentic-One uses a central Orchestrator to plan, delegate tasks to specialized agents, track progress, and initiate replanning~\cite{fourney2024magentic}. These systems exemplify conversation-centric, workflow-based, and orchestrator-centered coordination, respectively, but all produce dependency-structured trajectories in which earlier decisions shape subsequent messages, actions, and observations. Consequently, MAS debugging and repair must examine how failures emerge and propagate through the execution trajectory rather than relying solely on the terminal output.

\subsection{MAS Debugging and Repair}

MAS debugging aims to make multi-agent executions inspectable by identifying failure types, responsible components, and error locations. Existing work develops structured observability mechanisms, failure taxonomies, annotated traces, and localization benchmarks for agentic workflows~\cite{dong2024agentops,alsayyad2026agenttrace,shah2026characterizing,cemri2025multi,deshpande2025trail,zhang2025agent}. These studies provide foundations for describing and locating failures, but their traces are primarily used for post-hoc analysis. In contrast, \sys records model and tool interactions together with their dependencies, execution order, and realized results, turning execution traces into replayable records that support verifiable reconstruction.

MAS repair commonly regenerates failed behavior using self-generated feedback, peer critique, tool-grounded evidence, runtime diagnosis, or dependency-aware localization~\cite{madaan2023self,shinn2023reflexion,du2023improving,gou2024critic,zhang2026efficient,zhao2026debugging,nanda2026wink,zhao2026agenttether}. However, these methods are generally evaluated by terminal success without consistently controlling the execution before repair. Consequently, a successful rerun may repair the recorded failure or merely avoid it through stochastic resampling. Building on execution replay for controlling nondeterminism~\cite{ronsse2000executionreplaydebugging}, \sys reconstructs the observed execution before an evidence-supported intervention, enabling our study to distinguish targeted repair from stochastic repair.

\section{System Design}
\label{sec:system-design}

\begin{figure*}[t]
    \centering
    \paperfigure[
        width=\textwidth,
        keepaspectratio
    ]{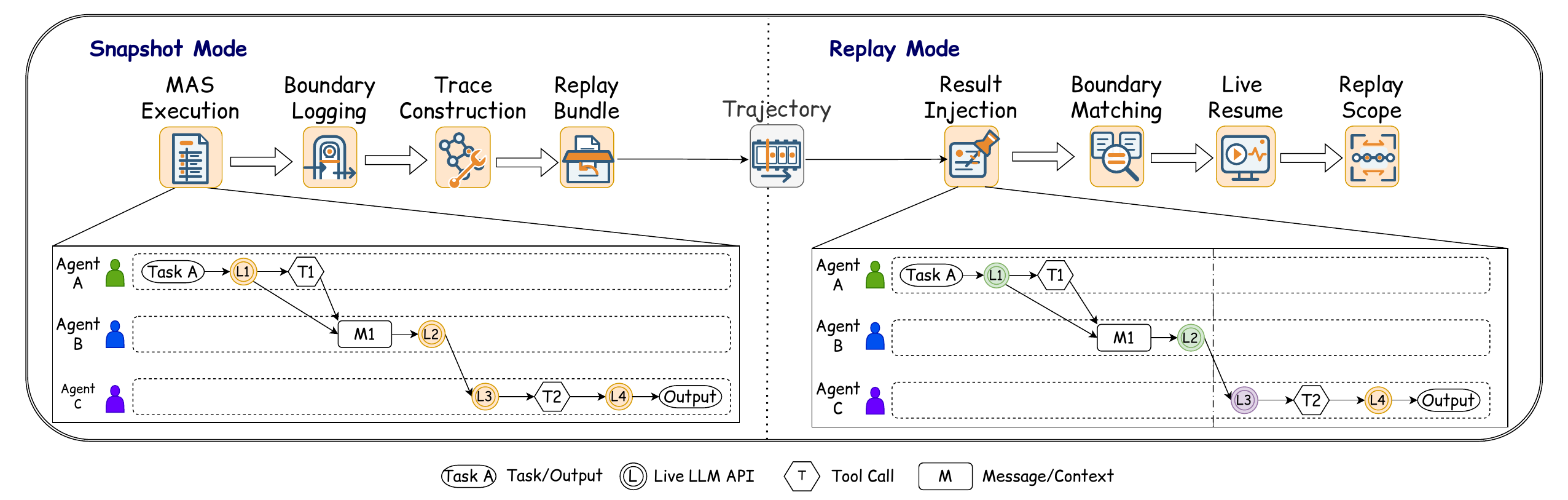}
    \caption{Overview of the \sys workflow.
    \scriptsize{(Snapshot Mode transforms a native MAS execution into a recorded trajectory through Boundary Logging, Trace Construction, and Replay Bundle. Replay Mode consumes the trajectory, reconstructs the target prefix through Result Injection and Boundary Matching, applies the intervention through Live Resume, and specifies the resulting guarantee through Replay Scope. The lower panels provide an illustrative source and replayed execution.)}}
    \label{fig:logging-replay}
    \Description{The SymTrace snapshot and replay workflow, from boundary logging and trace construction to validated prefix replay and live intervention.}
\end{figure*}

Figure~\ref{fig:logging-replay} presents the two-mode workflow of \sys. The pipeline takes a native MAS execution, including its task, represented initial state, runtime configuration, and realized model and tool interactions, as input. Snapshot Mode processes this execution through Boundary Logging and Trace Construction, and then uses Replay Bundle to materialize the recorded trajectory. Replay Mode takes this trajectory together with an intervention target as input, reconstructs the target prefix through Result Injection and Boundary Matching, and applies the intervention through Live Resume. Replay Scope finally specifies the guarantees attached to the reconstructed prefix and the newly generated suffix. The overall output is a new MAS trajectory that preserves the validated execution before the target and may diverge from the source trajectory from the intervention onward.

\subsection{Snapshot Mode}
\label{sec:snapshot-mode}

Snapshot Mode transforms a complete native MAS execution into a structured trajectory that contains the information required for subsequent replay. Specifically, Boundary Logging records the realized model and tool interactions, Trace Construction organizes the recorded events into their dependency structure and observed order, and Replay Bundle combines these results with the task and runtime configuration. The resulting bundle is the output of Snapshot Mode and the trajectory input to Replay Mode.

\noindent{\bf Boundary Logging.}
During the native execution, framework-specific hooks intercept exposed LLM request-response pairs and tool call-observation pairs without changing the MAS scheduler, agent logic, or state-update procedures. Each request continues to its live endpoint, while \sys records the realized request, result, and event position. This process converts the runtime interactions into ordered boundary records $R$, which preserve the nondeterministic model and tool values needed for replay.

\noindent{\bf Trace Construction.}
The boundary records describe the values observed at individual model and tool interactions but do not capture how those interactions depend on one another. Trace Construction therefore organizes the realized events into an event-dependency graph $G=(V,E)$, whose edges represent data or control dependencies. Repeated workflow iterations are stored as distinct event instances, and the observed event order $O=(v_1,\ldots,v_n)$ is recorded separately because the graph may permit multiple valid schedules. The resulting $G$ and $O$ provide the structural information associated with the boundary values in $R$.

\noindent{\bf Replay Bundle.}
Replay Bundle combines the recorded values and execution structure with the information required to restart the native MAS:
\begin{equation}
\mathcal{S}
=
\left\langle
T,s_0,c,G,O,R
\right\rangle,
\label{eq:replay-bundle}
\end{equation}
where $T$ is the task, $s_0$ is the represented initial state, and $c$ is the runtime configuration. The resulting bundle $\mathcal{S}$ materializes the recorded trajectory transferred from Snapshot Mode to Replay Mode.

\subsection{Replay Mode}
\label{sec:replay-mode}

Replay Mode reconstructs the recorded execution before a designated target and resumes the MAS from that point under an intervention. Specifically, Result Injection associates each intercepted prefix call with its historical boundary result, Boundary Matching verifies that the current call corresponds to the expected record, and Live Resume applies the intervention and restores live execution. Replay Scope then characterizes which part of the resulting trajectory is guaranteed by replay. We refer to this procedure as selective replay because only the recorded prefix before the target is reconstructed, while execution from the target onward remains live.

\noindent{\bf Result Injection.}
Replay Mode begins by restarting the native MAS using $T$, $s_0$, and $c$ from the replay bundle. Whenever the restarted MAS reaches a model or tool boundary before the target, Result Injection retrieves the corresponding historical record from $R$ instead of immediately invoking the live endpoint. The intercepted call and candidate record are then supplied to Boundary Matching, so the recorded result is returned only after the correspondence has been verified.

\noindent{\bf Boundary Matching.}
Boundary Matching validates the intercepted call against the expected record using its event position and canonicalized request content. A mismatch terminates replay, whereas a successful match authorizes the recorded result to be returned to the native MAS. Processing this result advances the MAS to its next represented state and produces the next boundary call. Repeating this process according to $G$ and $O$ reconstructs the target prefix, while content hashes validate the reused prefix nodes.

\noindent{\bf Live Resume.}
Once the verified prefix reaches the designated target, Live Resume stops returning historical boundary results and applies the intervention. Model and tool interactions are restored to their live endpoints, while the native scheduler and state-update procedures continue unchanged. This stage uses the reconstructed prefix and intervention as its starting state and produces a newly generated downstream trajectory.

\noindent{\bf Replay Scope.}
Replay Scope defines which parts of the resulting trajectory are reconstructed from recorded results and which parts remain live. Before the target, every reused boundary result is returned only after strict matching, preserving the represented logical history of the recorded execution. From the intervention onward, model generation, tool responses, and downstream decisions remain live and may differ from the source trajectory. This guarantee holds when native transitions are deterministic under the recorded boundary values and relevant external state is reset, isolated, or restored. It does not require equality of inaccessible framework internals or deterministic behavior in the new suffix.

\section{Dataset Construction}
\label{sec:dataset-construction}

Based on WebArena-Verified Hard~\cite{zhou2024webarena} and AssistantBench~\cite{yoran2024assistantbench}, we construct and annotate \ds, comprising 536 evaluator-confirmed failure trajectories with structured execution traces, event graphs, localized failure nodes, and multi-label annotations.

\subsection{Candidate Selection}
\label{sec:task-pool-construction}

The candidate set comprises all 258 WebArena-Verified Hard tasks and the 33 eligible AssistantBench development tasks with an available task description, reference answer, and gold URL. We retain all 33 AssistantBench tasks and the first 167 WebArena-Verified Hard tasks in their published order, producing a deterministic pool of 200 tasks before any MAS execution or repair experiment. The final pool contains 33 Web-QA, 70 mutation, 25 navigation, and 72 retrieval tasks. We select these benchmarks because they provide multi-step Web tasks with externally verifiable outcomes.

\subsection{Trajectory Collection}
\label{sec:failure-trajectory-collection}

Each task is executed once with AG2~\cite{wu2023autogen}, CrewAI~\cite{crewai2026}, and Magentic-One~\cite{fourney2024magentic}, producing $200\times3=600$ initial task-MAS executions. The native benchmark evaluators identify 536 failures with complete trace and graph artifacts, while the remaining 64 executions are excluded before annotation. The 501 WebArena-Verified executions contribute 462 failures and the 99 AssistantBench executions contribute 74 failures. The retained trajectories comprise 171 AG2, 184 CrewAI, and 181 Magentic-One executions.

\subsection{Failure Annotation}
\label{sec:annotation-annotation}

We derive C1-C4 from the Multi-Agent System Failure Taxonomy~\cite{cemri2025multi}, which identifies 14 fine-grained failure modes under three broader categories. We retain the four most prevalent consolidated patterns in the source analysis, producing the non-exclusive, trace-observable categories in Table~\ref{tab:repair-signals}. 

All four annotators have software-engineering backgrounds and experience in software development and code analysis. Before the main annotation, they calibrate on reference-labeled examples and practice trajectories using shared category, evidence, and node-selection guidance. Three annotators then independently assign all applicable categories, select one primary category, and identify the earliest trace-supported actionable node with evidence and a rationale. A fourth annotator reviews the task, complete trace, event graph, and initial annotations to produce the final annotation, with authority to revise any decision rather than applying majority vote or label union.

The final category set differs from the initial label union in 113 of 536 trajectories (21.08\%). Agreement is measured only among the three independent annotators: Fleiss' $\kappa$ is 0.62 for the primary category and 0.81 for failure-node type, while exact node agreement is 73.88\% among all three annotators and 95.90\% for at least two annotators. All $536\times(3+1)=2{,}144$ initial and final node selections exist in their source event graphs and match their recorded node types. The release includes the task manifest, linked traces and graphs, initial and final annotations, rationales, and supporting evidence. Appendix~B provides the complete annotation guidance and reliability analysis.

\begin{table}[t]
\centering
\small
\setlength{\tabcolsep}{3pt}
\caption{Diagnosis-to-repair mapping. \scriptsize{(Categories represent trace-localized intervention signals rather than task-level failure verdicts.)}}
\label{tab:repair-signals}
\begin{tabular}{
    >{\centering\arraybackslash}m{0.19\columnwidth}
    >{\raggedright\arraybackslash}p{0.73\columnwidth}}
\toprule
\textbf{Category} & \textbf{Repair signal} \\
\midrule
$C_1$
& Implicated constraint and conflicting behavior; request a compliant alternative. \\
$C_2$
& Repeated history and unfinished objective; request an action that makes progress. \\
$C_3$
& Unresolved runtime condition and its evidence; require resolution before continuing. \\
$C_4$
& Explicit plan, action, and outcome; request a decision that resolves their inconsistency. \\
\bottomrule
\end{tabular}
\end{table}

\section{Study}

\subsection{Setup}
\label{sec:experimental-setup}

\noindent{\bf Study scope.}
The primary evaluation applies AG2~\cite{wu2023autogen}, CrewAI (CA)~\cite{crewai2026}, and Magentic-One (MO)~\cite{fourney2024magentic} to the 200 fixed tasks used to construct \ds, yielding 536 source failures from 600 runs (Section~\ref{sec:dataset-construction}). Each source failure is one experimental unit; executions of the same task by different MASs are treated separately. The evaluation set is frozen before replay and repair.

\noindent{\bf Baselines.}
We compare three representative task-level debugging and repair methods. For each source failure, all methods use the same task input, MAS, model alias, temperature, and native evaluator.
\noindent\ding{182} \textit{Unguided Full Rerun} is a retry-based repair method that restarts the complete MAS execution from the original task without using information from the failed execution~\cite{brown2024large}.
\noindent\ding{183} \textit{Self-Reflection} is a feedback-based repair method that provides the MAS with its failed final output and asks it to identify possible mistakes before solving the task again~\cite{madaan2023self,shinn2023reflexion}.
\noindent\ding{184} \textit{Critic-Agent} is a critic-based repair method that employs a separate critic agent to review the failed output and provide corrective guidance for a new complete execution~\cite{gou2024critic}.
Each task-level method receives up to three complete attempts and stops after its first success, whereas each node-level method receives one selective-replay intervention.

\noindent{\bf Evaluation metrics.}
Following $\tau$-bench~\cite{yao2024tau} and the $pass@k$ convention~\cite{chen2021evaluating}, we define two metrics parameterized by the attempt count $k$. Let $N$ denote the number of source cases, $z_{i,a}^{m}$ indicate whether method $m$ reproduces the source failure for case $i$ in attempt $a$, and $y_{i,a}^{m}$ indicate whether the native evaluator accepts the corresponding repair.

\begin{itemize}
    \item \textbf{Failure reproduction} ($\mathrm{rep}_{k}$). Given three executions per case, let $\mathcal{A}_{k}=\{S\subseteq\{1,2,3\}:|S|=k\}$ denote all subsets of $k$ executions:
    \begin{equation}
        \mathrm{rep}_{k}(m)
        =
        \frac{1}{N|\mathcal{A}_{k}|}
        \sum_{i=1}^{N}
        \sum_{S\in\mathcal{A}_{k}}
        \prod_{a\in S}z_{i,a}^{m}.
        \label{eq:evaluation-metrics}
    \end{equation}
    This metric describes the proportion of execution subsets in which all $k$ executions reproduce the source failure.

    \item \textbf{Repair success} ($\mathrm{pass@}k$):
    \begin{equation}
        \mathrm{pass@}k(m)
        =
        \frac{1}{N}
        \sum_{i=1}^{N}
        \mathbf{1}\!\left[
        \max_{a\in\{1,\ldots,k\}}y_{i,a}^{m}=1
        \right].
    \end{equation}
    This metric describes the proportion of source failures that receive at least one evaluator-accepted repair within $k$ attempts.
\end{itemize}

\noindent{\bf Model and efficiency reporting.}
All experimental conditions use \path{deepseek-v4-flash}~\cite{xu2026deepseek} at temperature 0.00 through the same OpenAI-compatible endpoint. We report captured API calls and accepted repairs per 1,000 calls as secondary efficiency measures, compared primarily within each MAS.

\noindent{\bf Statistical analysis.}
Repair rates are reported with 95\% Wilson confidence intervals to quantify uncertainty in the estimated proportions. Paired rate differences are reported with 95\% confidence intervals obtained from 10,000 case-level bootstrap resamples, measuring the magnitude and uncertainty of the difference between methods. Two-sided exact McNemar tests compare paired binary outcomes on the same source cases, while Holm correction controls the family-wise error rate across prespecified multiple comparisons within each evaluation setting and MAS. Appendix~C provides the complete execution and statistical conventions.

\subsection{RQ1: How Reliably Can MAS Failures Be Reproduced?}
\label{sec:rq1}

\noindent{\bf Experimental design.}
To assess whether reconstructing the recorded execution history reproduces failures more reliably than resampling an entire execution, we analyze 536 source failures with three attempts per method, comparing \textit{\sys Replay}, which reconstructs the recorded prefix and resumes at the target failure node, against \textit{Unguided Full Rerun}, which restarts the original task without using any guidance. We use an LLM-as-a-judge pipeline adapted to our taxonomy from \citet{cemri2025multi}, with the \ds labels as references. Table~\ref{tab:rq1-reproduction} reports the results, where the rows represent the two methods, the columns represent the three MASs and their aggregate, and the two column groups report $\mathrm{rep}_{1}$ and $\mathrm{rep}_{3}$, respectively.

\begin{table}[t]
\caption{Same-failure reproduction rates (\%) by MAS on the 536 \ds source failures. \scriptsize{(CA and MO denote CrewAI and Magentic-One; Rerun and Replay denote Unguided Full Rerun and \sys Replay. Bold marks the higher value in each column.)}}
\label{tab:rq1-reproduction}
\centering
\footnotesize
\setlength{\tabcolsep}{1.6pt}
\resizebox{\columnwidth}{!}{%
\begin{tabular}{lcccccccc}
\toprule
& \multicolumn{4}{c}{\textbf{\boldmath$\mathrm{rep}_{1}$ (\%)}}
& \multicolumn{4}{c}{\textbf{\boldmath$\mathrm{rep}_{3}$ (\%)}} \\
\cmidrule(lr){2-5}\cmidrule(lr){6-9}
\textbf{Method}
& \textbf{AG2} & \textbf{CA} & \textbf{MO} & \textbf{Total}
& \textbf{AG2} & \textbf{CA} & \textbf{MO} & \textbf{Total} \\
\midrule
Rerun  & 64.13 & 77.36 & 62.06 & 67.97 & 36.26 & 52.72 & 34.81 & 41.42 \\
Replay & \textbf{80.70} & \textbf{81.88} & \textbf{79.74} & \textbf{80.78} & \textbf{50.29} & \textbf{53.26} & \textbf{53.59} & \textbf{52.43} \\
\bottomrule
\end{tabular}
}
\end{table}

\noindent{\bf Analysis of reproduction reliability.}
The reproduction results show that \sys achieves a consistent advantage over full rerun across all three MASs. All reproduced prefix nodes pass content-hash validation, yielding 100.00\% prefix exactness. The remaining gap between exact prefix reconstruction and end-to-end reproduction can be attributed to short replay prefixes, post-target execution variation, and evaluation uncertainty.

First, most cases preserve only a short execution prefix. Specifically, 381 of the 536 cases (71.08\%) reuse only 2-3 nodes, including 738 of the 1,608 attempts (45.90\%) that reuse exactly two. Within this dominant group, the $\mathrm{rep}_{1}$ and $\mathrm{rep}_{3}$ gains over full rerun are 9.80 and 6.82 percentage points. The corresponding gains increase to 17.76 and 18.69 points for 4-8 reused nodes and to 25.69 and 27.08 points for at least nine nodes.

Second, \sys reuses only the nodes preceding the failure target. The target and its successors remain subject to live model and tool execution. For \texttt{webarena\_verified\_hard\_127} on AG2, three attempts reconstruct identical prefixes but generate three different unsupported answers without intervening tool calls. For \texttt{webarena\_verified\_hard\_267}, the source execution receives an HTTP 200 response from the Wikipedia API, whereas replay receives an HTTP 429 response. Such executions may retain the same high-level failure category while exhibiting different case-specific failures and are therefore not counted as successful reproductions.

Third, LLM-as-a-judge errors introduce measurement uncertainty. In a stratified secondary audit of 72 judgments, 64 agree with the original labels, five are confirmed disagreements, and three remain uncertain. After weighting by stratum, the estimated confirmed-disagreement rate is 3.30\%, the uncertainty rate is 3.67\%, and their combined sensitivity upper bound is 6.96\%. All five confirmed disagreements are false negatives in which a genuine reproduction was labeled as different\_failure.

\finding{\sys improves failure reproduction by exactly reconstructing the execution before the failure target. Although it cannot control model and environmental variation after that target, its advantage grows when more preceding steps can be reused, making it particularly suitable for realistic long-horizon MAS tasks.}


\subsection{RQ2: Can Task-Level Re-execution Reliably Repair Failed MAS Executions?}
\label{sec:rq2}

\noindent{\bf Experimental design.}
To understand whether task-level repair methods correct source failures or mainly benefit from repeated sampling, we use \ds as the evaluation basis and apply three baseline repair methods to its 536 failed executions from AG2, CrewAI, and Magentic-One. The results are presented in Table~\ref{tab:rq2-three-systems}. Each row reports one subset of the 536 failures, grouped in the first column by MAS and by primary failure category (C1-C4), with its case count and the $\mathrm{pass@3}$ repair rate of each baseline. Both groupings share the same overall total.

\begin{table}[t]
\caption{Task-level $\mathrm{pass@3}$ results (\%) of the three task-level baselines on \ds. \scriptsize{(The first column groups the same 536 failures by MAS and by primary failure category (C1-C4). Both breakdowns share the single overall \textbf{Total}. `Rerun', `Self', and `Critic' denote Unguided Full Rerun, Self-Reflection, and Critic-Agent.)}}
\label{tab:rq2-three-systems}
\centering
\footnotesize
\setlength{\tabcolsep}{4pt}
\resizebox{\columnwidth}{!}{%
\begin{tabular}{llrrrr}
\toprule
 & \multirow{2}{*}{\textbf{Group}} & \multirow{2}{*}{\textbf{Cases}} & \multicolumn{3}{c}{\textbf{\boldmath$\mathrm{pass@3}$ (\%)}} \\
\cmidrule(lr){4-6}
 & & & \textbf{Rerun} & \textbf{Self} & \textbf{Critic} \\
\midrule
\multirow{3}{*}{\textbf{MAS}}
 & AG2          & 171 & 8.19 & 4.68 & 4.09 \\
 & CrewAI       & 184 & 3.80 & 2.17 & 2.72 \\
 & Magentic-One & 181 & 8.84 & 6.08 & 4.42 \\
\midrule
\multirow{4}{*}{\textbf{Category}}
 & C1 & 125 & 9.60 & 4.80 & 2.40 \\
 & C2 &  69 & 4.35 & 2.90 & 5.80 \\
 & C3 & 203 & 5.42 & 3.45 & 1.97 \\
 & C4 & 139 & 7.91 & 5.76 & 6.47 \\
\midrule
\multicolumn{2}{l}{\textbf{Total}} & \textbf{536} & \textbf{6.90} & \textbf{4.29} & \textbf{3.73} \\
\bottomrule
\end{tabular}
}
\end{table}

\noindent{\bf Analysis across evaluation settings.}
The results on \ds show that generic reflection and critique do not provide a consistent repair advantage over unguided task restart. Unguided Full Rerun performs best for every MAS, while the category breakdown provides no evidence that the feedback-based methods systematically address particular failure types. This suggests that their additional feedback does not reliably target the source failure mechanism.

The comparison with the original CRITIC and Reflexion studies highlights two factors affecting reported repair rates: task difficulty and repeated sampling. \ds consists of failed long-horizon Web executions involving multi-agent coordination, dynamic information acquisition, and multiple dependent actions, making successful regeneration particularly difficult. More importantly, both original studies report increasing success over successive trials~\cite{gou2024critic,shinn2023reflexion}. Because each additional trial provides another opportunity to sample a different answer or trajectory, the resulting gains combine feedback effects with repeated regeneration. Under the same three-attempt budget on \ds, the feedback-based methods still do not outperform unguided rerunning.

\noindent{\bf Analysis of outcome instability.}
Complete re-execution changes outcomes in both directions, confirming the stochasticity of the task-level baselines and showing that an accepted rerun does not necessarily indicate repair of the source failure. To examine this behavior, we use 54 initially successful executions recorded by \sys during the same data-collection process and rerun each execution three times under the same runtime configuration. Among the resulting 162 attempts, 85 result in failures, and 39 of the 54 source cases regress at least once.

\noindent{\bf Case analysis of repair mechanisms.}
A CrewAI case shows that an evaluator-accepted answer can be produced without resolving the source failure mechanism. The task asks for the rate of snowy New Year's Eves in Chicago from 2014 to 2023, but the source execution and all three repair methods lack the required observations because the extracted dynamic pages do not contain them. Self-Reflection and Critic-Agent report that the requested rate cannot be determined, whereas Unguided Full Rerun returns the accepted answer of 30.00\% while acknowledging that it relies on general historical patterns rather than the required evidence.

\finding{Task-level methods repair primarily through stochastic regeneration rather than failure-specific repair. Re-execution can both repair failed outcomes and destabilize successful ones, while feedback-based variants provide no advantage under equal budgets. Thus, observed repair alone does not demonstrate that the source failure was identified and repaired.}

\subsection{RQ3: Are Node-Level Symptom Signals Actionable for Targeted Repair?}
\label{sec:rq3}

\noindent{\bf Experimental design.}
To evaluate whether trace-localized symptoms provide actionable evidence for targeted repair, we analyze the 536 source failures from AG2, CrewAI (CA), and Magentic-One (MO). We introduce \textit{Suspicious-Node Intervention}, a node-level repair method that treats a trace node as a candidate intervention anchor when its local behavior exhibits observable failure symptoms connected to the unsuccessful outcome. During execution, deterministic rules and a semantic judge jointly evaluate each newly completed node for the C1-C4 symptoms defined in Table~\ref{tab:repair-signals}. When the judge assigns the current node a suspicion score above a predefined threshold, the controller suspends the ongoing MAS execution. Because the suspicious behavior has already occurred by the time it is detected, \sys reconstructs the execution prefix leading to that boundary, applies a repair instruction conditioned on the judge's C1-C4 classification, and then resumes live execution from the intervention point. We compare this method with Random-Node Intervention and Last-Node Intervention, which receive the same single selective-replay opportunity but no symptom evidence, and with the three task-level debugging and repair methods defined in Section~\ref{sec:rq2}.

Table~\ref{tab:rq3-repair-utility} reports the budgeted pass rate of each method on each MAS and over all 536 failures. For the three node-level methods, the reported value is $\mathrm{pass@1}$ because each method receives one selective-replay intervention. For the task-level methods, it is $\mathrm{pass@3}$ because they receive up to three complete-execution attempts and stop after the first success. This comparison is conservative with respect to Suspicious-Node Intervention: the proposed method must repair the task through a single symptom-conditioned intervention, whereas the task-level baselines have three opportunities to obtain a successful outcome through complete re-execution and stochastic resampling. Thus, the larger pass rate of Suspicious-Node Intervention cannot be attributed to a larger attempt budget. Appendix~F presents the complete pipeline, detection rules, threshold configuration, and intervention prompts.

\begin{table}[t]
\caption{RQ3 repair utility on the 536 \ds source failures. \scriptsize{(AG2, CrewAI (CA), and Magentic-One (MO) contribute 171, 184, and 181 failures, respectively. `Att.' is the attempt budget: node-level methods use one selective-replay intervention and task-level methods use up to three complete attempts. Bold marks the best result.)}}
\label{tab:rq3-repair-utility}
\centering
\footnotesize
\setlength{\tabcolsep}{5pt}
\renewcommand{\arraystretch}{1.08}
\resizebox{\columnwidth}{!}{%
\begin{tabular}{lcrrrr}
\toprule
\multirow{2}{*}{\textbf{Method}} & \multirow{2}{*}{\textbf{Att.}} & \multicolumn{4}{c}{\textbf{Pass rate (\%)}} \\
\cmidrule(lr){3-6}
 & & \textbf{AG2} & \textbf{CA} & \textbf{MO} & \textbf{Total} \\
\midrule
Last-Node           & 1 & 0.58 & 1.63 & 1.66 & 1.31 \\
Random-Node         & 1 & 2.34 & 4.89 & 3.87 & 3.73 \\
Critic-Agent        & 3 & 4.09 & 2.72 & 4.42 & 3.73 \\
Self-Reflection     & 3 & 4.68 & 2.17 & 6.08 & 4.29 \\
Unguided Full Rerun & 3 & 8.19 & 3.80 & 8.84 & 6.90 \\
\textbf{Suspicious-Node} & 1 & \textbf{16.37} & \textbf{25.00} & \textbf{18.78} & \textbf{20.15} \\
\bottomrule
\end{tabular}
}
\end{table}

\noindent{\bf Analysis of repair effectiveness.}
Suspicious-Node Intervention achieves the highest overall repair rate and nearly triples the strongest task-level baseline, despite receiving only one intervention rather than three complete attempts. It also substantially outperforms Random-Node Intervention and Last-Node Intervention under the same selective-replay budget. The comparison with Random-Node Intervention and Last-Node Intervention shows that selective replay alone does not explain the improvement. Together, these results indicate that the improvement is not explained solely by additional sampling opportunities or access to an intervention point. Suspicious-Node Intervention additionally uses symptom-based target selection and repair guidance. Because target selection and repair guidance are evaluated jointly, the experiment does not attribute the full improvement to either component in isolation.

\noindent{\bf Analysis of cross-MAS consistency.}
Although absolute repair effectiveness varies across architectures, the method ranking remains stable: Suspicious-Node Intervention performs best on every MAS. This consistency is notable because the task-level baselines receive up to three opportunities to benefit from stochastic resampling, whereas Suspicious-Node Intervention uses only one symptom-guided intervention. It significantly outperforms Unguided Full Rerun on all three MASs after within-MAS Holm correction, showing that the overall advantage is not driven by a single architecture. Localized symptom evidence therefore provides a more consistent repair signal than repeated full-trajectory regeneration.

\finding{Across MAS architectures, repair effectiveness depends more on precise execution control than on repeated regeneration. Localized symptoms provide actionable repair guidance, whereas full-trajectory resampling cannot reveal what was actually corrected.}

\section{Discussion}

This section discusses the limitations and potential future directions.
\ding{182} {\bf Data Coverage}. We ground \ds in established, externally verifiable benchmarks (i.e., WebArena-Verified Hard, AssistantBench).
This yields reproducible, outcome-verified failures, but benchmark-derived tasks cannot capture the full diversity of failures in deployed MASs.
Therefore, collecting real-world failure trajectories and covering further MAS structures, interaction patterns, and task domains is a valuable future direction.
\ding{183} {\bf Automated Evaluation}.
In the failure-reproduction experiment, we use LLM judges to determine whether a generated execution reproduces the source failure. The judges localize and classify the failure in the generated trajectory, compare these results with the reference location and category, and decide whether they represent the same failure. Although this automated evaluation enables large-scale repeated experiments, LLM judges may misinterpret long trajectories, ambiguous evidence, or semantically similar failure locations, and our manual spot checks identify occasional errors. Future work can incorporate expert verification for disagreements and low-confidence cases, develop hybrid human-LLM annotation, and report human-LLM agreement alongside automated results.
\ding{184} {\bf Model Coverage}.
We use the same recent model alias and generation configuration across all MASs and repair methods. This controlled setting is necessary for our large-scale paired comparisons: it prevents differences in model capability or version from confounding differences between MAS architectures and repair strategies, while evaluating \sys with a model representative of current agent systems. Nevertheless, results from a single model may not generalize to other model families, providers, scales, or future versions. Future studies can repeat the evaluation across a broader set of models to test whether the observed reproduction and repair patterns remain consistent.

\section{Conclusion}

This study distinguishes correction of a recorded MAS failure from stochastic repair through a different execution. We introduce \sys as a controlled evaluation framework, construct \ds with human annotation, and conduct a cross-system study of failure reproduction and repair. Across systems and tasks, failure identity depends on execution history rather than task input alone. Full task regeneration conflates repair with outcome variability, making terminal success insufficient evidence that the original mechanism was corrected. Localized symptoms can nevertheless guide effective intervention without uniquely identifying a root cause. Reliable repair evaluation should therefore preserve the history before intervention, localize the intended change, and assess whether the failure mechanism was corrected.

\bibliographystyle{ACM-Reference-Format}
\bibliography{ref}

\clearpage
\input{appendix}

\end{document}

%% file: appendix.tex

\definecolor{MASDNavy}{HTML}{17324D}
\definecolor{MASDBlue}{HTML}{0072B2}
\definecolor{MASDAmber}{HTML}{D89000}
\definecolor{MASDRed}{HTML}{D55E00}
\definecolor{MASDPurple}{HTML}{7A5195}
\definecolor{MASDGreen}{HTML}{218739}
\definecolor{MASDInk}{HTML}{1F2933}
\definecolor{MASDGray}{HTML}{657786}
\definecolor{MASDLine}{HTML}{CBD5E1}
\definecolor{MASDFog}{HTML}{F4F7FA}

\newcommand{\masdcheck}{\textcolor{MASDGreen}{\ding{51}}}
\newcommand{\masdwarning}{\textcolor{MASDAmber}{\ding{115}}}

\newcommand{\masdpill}[2]{%
  \tikz[baseline=(masdpillnode.base)]{
    \node[
      anchor=base,
      rounded corners=0.8mm,
      inner xsep=1.1mm,
      inner ysep=0.35mm,
      fill=#1,
      text=white,
      font=\sffamily\bfseries\scriptsize
    ] (masdpillnode) {\strut #2};
  }%
}

\lstdefinestyle{MASDTrace}{
  basicstyle=\ttfamily\fontsize{7.1}{8.5}\selectfont,
  columns=fullflexible,
  keepspaces=true,
  breaklines=true,
  breakatwhitespace=false,
  showstringspaces=false,
  numbers=left,
  numberstyle=\tiny\color{MASDGray},
  numbersep=5pt,
  frame=single,
  framerule=0.45pt,
  rulecolor=\color{MASDLine},
  backgroundcolor=\color{MASDFog},
  xleftmargin=1.75em,
  framexleftmargin=1.4em,
  aboveskip=0.5em,
  belowskip=0.25em
}

\newtcolorbox{masdgold}[4][]{%
  enhanced,
  breakable,
  colback=#4!3!white,
  colframe=#4,
  colbacktitle=#4,
  coltitle=white,
  boxrule=0.85pt,
  borderline west={2.4pt}{0pt}{#4},
  arc=1.8mm,
  left=1.6mm,
  right=1.6mm,
  top=1.4mm,
  bottom=1.4mm,
  before skip=8pt,
  after skip=10pt,
  fontupper=\small,
  before upper=\raggedright,
  fonttitle=\sffamily\bfseries,
  title={%
    \large #2\quad
    \normalsize #3
    \hfill
    {\scriptsize\ding{51}\ TRACE-VERIFIED}%
  },
  #1
}

\newtcolorbox{masdpanel}[2][]{%
  enhanced,
  colback=white,
  colframe=MASDLine,
  boxrule=0.45pt,
  arc=1.2mm,
  left=1.3mm,
  right=1.3mm,
  top=1.1mm,
  bottom=1.1mm,
  before skip=5pt,
  after skip=5pt,
  fonttitle=\sffamily\bfseries\small,
  coltitle=MASDNavy,
  colbacktitle=MASDFog,
  title={#2},
  #1
}

\newtcolorbox{masdverdict}[2]{%
  enhanced,
  colback=#1!7!white,
  colframe=#1,
  boxrule=0.75pt,
  borderline west={1.8pt}{0pt}{#1},
  arc=1.2mm,
  left=1.4mm,
  right=1.4mm,
  top=1.2mm,
  bottom=1.2mm,
  before skip=5pt,
  after skip=5pt,
  fonttitle=\sffamily\bfseries\small,
  coltitle=white,
  colbacktitle=#1,
  before upper=\raggedright,
  title={#2}
}

\newtcolorbox{masdreview}{%
  enhanced,
  colback=MASDNavy!4!white,
  colframe=MASDNavy!60,
  boxrule=0.55pt,
  arc=1.2mm,
  left=1.4mm,
  right=1.4mm,
  top=1.0mm,
  bottom=1.0mm,
  before skip=5pt,
  after skip=2pt,
  fonttitle=\sffamily\bfseries\small,
  coltitle=MASDNavy,
  colbacktitle=MASDNavy!9!white,
  title={Reviewer takeaway}
}

\newcommand{\masdmeta}[7]{%
  \begin{tabularx}{\linewidth}{
    @{}>{\raggedright\arraybackslash\bfseries\color{MASDNavy}}p{0.22\linewidth}
    >{\raggedright\arraybackslash}X@{}
  }
    Case / annotation
      & \path{#1} / \path{#2} \\[1pt]
    Task ID
      & \path{#3} \\[1pt]
    Source / MAS
      & #4 \\[1pt]
    Trace size
      & #5 \\[1pt]
    Gold failure node
      & \path{#6} \\[1pt]
    Raw-log bundle
      & \path{#7} \\
  \end{tabularx}
}

\tikzset{
  masdstep/.style={
    rounded corners=1.1mm,
    draw=MASDNavy!70,
    fill=MASDFog,
    line width=0.55pt,
    minimum height=8.5mm,
    text width=1.82cm,
    align=center,
    inner sep=1.1mm,
    font=\sffamily\scriptsize,
    text=MASDInk
  },
  masdok/.style={
    masdstep,
    draw=MASDGreen,
    fill=MASDGreen!7!white
  },
  masdfail/.style={
    masdstep,
    draw=MASDRed,
    fill=MASDRed!8!white,
    line width=0.9pt
  },
  masdambiguous/.style={
    masdstep,
    draw=MASDPurple,
    fill=MASDPurple!8!white,
    line width=0.9pt
  },
  masdloop/.style={
    masdstep,
    draw=MASDAmber,
    fill=MASDAmber!10!white,
    line width=0.9pt
  },
  masdarrow/.style={
    -{Latex[length=1.8mm]},
    line width=0.6pt,
    draw=MASDGray
  }
}

\appendix
\setcounter{secnumdepth}{2}

\section{Replay Matching Rules and Runtime Assumptions}
\label{app:replay-details}

\subsection{Observed Execution Order}

The event-dependency graph $G=(V,E)$ defines a partial order but does not uniquely determine a concurrent execution. Consistent with the main paper, \sys therefore records the observed event order
\begin{equation}
O=(v_1,\ldots,v_n),
\end{equation}
subject to
\begin{equation}
(v_i,v_j)\in E
\Longrightarrow
i<j.
\end{equation}
Unlike a topological ordering computed after execution, $O$ is the particular linear extension observed during the recorded run. It fixes the order in which state-affecting operations crossed an instrumented boundary. During replay, the recorded order is used solely for event matching and validation; the adapter does not replace or control the native MAS scheduler.

Algorithm~\ref{alg:snapshot-recording} records exposed events and boundary results without replacing native MAS control flow.

\subsection{Boundary Matching and Record Validation}

The ordered boundary records $R$ in the replay bundle preserve complete interactions. An LLM entry contains its prompt and ordered messages, model configuration, tool specification, expected response format, and returned response. A tool entry contains the tool identity, arguments, returned observation, and execution metadata. Each entry also retains its position in $O$.

For an intercepted boundary event $v_j$, the adapter validates the request against the expected record using its position in $O$ and canonicalized request content. A mismatch terminates replay at the first divergence, whereas a successful match authorizes the recorded result to be returned through the native boundary. After the result is materialized, its node content hash is compared with the source record to validate the reused prefix.

\begin{algorithm}[H]
\caption{Replay Bundle Recording}
\label{alg:snapshot-recording}
\footnotesize
\begin{algorithmic}[1]
\REQUIRE Task $T$, represented initial state $s_0$, runtime configuration $c$
\ENSURE Replay bundle $\mathcal{S}=\langle T,s_0,c,G,O,R\rangle$
\STATE Initialize $V,E\leftarrow\emptyset$ and $O,R\leftarrow()$
\STATE Start the native MAS with the framework adapter enabled
\WHILE{the native MAS has not terminated}
    \STATE Observe event $v$ and assign its identity, agent, type, and ordinal
    \STATE Add $v$ and its dependencies to $G=(V,E)$; append $v$ to $O$
    \IF{$v$ crosses an LLM or tool boundary}
        \STATE Capture request $x_v$ and configuration $\lambda_v$
        \STATE Execute the live boundary to obtain $y_v$
        \STATE Append $(v,x_v,\lambda_v,y_v)$ to $R$ and return $y_v$
    \ENDIF
\ENDWHILE
\STATE \textbf{return} $\mathcal{S}=\langle T,s_0,c,G,O,R\rangle$
\end{algorithmic}
\end{algorithm}

Algorithm~\ref{alg:replay} gives the fail-closed procedure for selective replay. A mismatch terminates at the first divergence rather than scanning forward for another record.

\begin{algorithm}[t]
\caption{Fail-Closed Selective Replay}
\label{alg:replay}
\small
\begin{algorithmic}[1]
\REQUIRE Replay bundle $\mathcal{S}=\langle T,s_0,c,G,O,R\rangle$
\REQUIRE Optional target $v_k$ and repair guidance $\Delta$
\ENSURE Replayed trace or first-divergence report
\STATE Restart the native MAS from $(T,s_0,c)$
\STATE $j\leftarrow1$; $\mathit{live}\leftarrow\mathrm{false}$
\WHILE{the MAS requests a model or tool boundary $v$}
    \IF{$\mathit{live}$}
        \STATE Execute $v$ live and return its result
    \ELSIF{$v=v_k$}
        \STATE Augment the request: $\widehat q_k\leftarrow q_k\oplus\Delta$
        \STATE $\mathit{live}\leftarrow\mathrm{true}$
        \STATE Execute the target boundary live using $\widehat q_k$
    \ELSE
        \IF{$j>|R|$ or the event position or canonicalized request content does not match $R_j$}
            \STATE \textbf{terminate} and report the first divergence
        \ENDIF
        \STATE Return the recorded result in $R_j$ through the native boundary
        \STATE Validate the materialized event position and content hash
        \STATE $j\leftarrow j+1$
    \ENDIF
\ENDWHILE
\STATE \textbf{return} the materialized execution trace
\end{algorithmic}
\end{algorithm}

In selective mode, Algorithm~\ref{alg:replay} stops before a designated LLM event $v_k$ and augments its boundary request with repair guidance $\Delta$,
\begin{equation}
\widehat{q}_k=q_k\oplus\Delta,
\end{equation}
before switching the target and suffix to live execution. The represented history before $v_k$ is preserved, while intentional divergence begins at the intervention boundary.

\subsection{Replay Conditions and Scope}

The replay guarantee relies on four conditions:

\begin{enumerate}
    \item the task, initialization, executable code, and runtime configuration are fixed;
    \item internal MAS transitions are deterministic when conditioned on the same represented state, input, and boundary result;
    \item every state-affecting model response, tool observation, external input, and scheduling decision is captured; and
    \item replay returns a recorded result only after validating the event position and canonicalized request content, and separately validates each materialized reused prefix node using its content hash.
\end{enumerate}

These conditions do not require the live model or external service itself to be deterministic. They make explicit the replay guarantee summarized in the main paper: selective replay reconstructs the represented logical execution prefix recorded in the replay bundle.

Although arbitrary hidden framework objects are not serialized, their represented logical values are reconstructed by native execution because they undergo the same transitions from the same initial state. The guarantee covers recorded boundary results, agent-visible messages, represented logical state, control decisions, and event order. It does not require identical memory addresses, physical thread interleavings, network latency, or wall-clock timing.

\subsection{External State Considerations}

A recorded tool result reconstructs the observation available to the MAS but does not reproduce a persistent side effect in an external service. Replaying a stored success response for a write operation, for example, does not recreate the written object in a live backend. Selective repair therefore requires that the live suffix not depend on an unreproduced mutation in the replayed prefix. This condition holds when the prefix contains no required persistent write, the environment is reset and the write can be safely re-executed, or a service-level checkpoint restores the required state. The reported evaluation includes all 536 source failures without filtering cases according to external-state dependence; when relevant external state is not reset, isolated, or restored, any resulting suffix variation lies outside the replay guarantee.

\section{Dataset Construction, Annotation, and Reliability}
\label{app:dataset-construction}

\subsection{Cohort Accounting}

Applying the three MASs to the 200 fixed tasks produces 600 initial task--MAS executions. The native benchmark evaluators identify 536 failed executions with complete trace and graph artifacts, all of which enter \ds and are subsequently annotated. The remaining 64 executions are evaluator-accepted and are excluded before annotation. The finalized dataset therefore contains 536 failures: 462 from WebArena-Verified and 74 from AssistantBench, comprising 171 AG2, 184 CrewAI, and 181 Magentic-One executions.

\subsection{Annotation Rules}

\noindent{\bf Multi-label taxonomy.}
The C1--C4 categories follow the diagnosis-to-repair mapping in the main paper. They are non-exclusive, trace-observable operational symptoms rather than mutually exclusive causal mechanisms, so annotators assign every category directly supported by the localized trace and separately select one primary category. The primary category is the best-supported intervention signal at the localized node; the guide imposes no category priority and does not treat it as a unique root cause. Dataset-level failure composition is computed from the complete category sets.

The four symptom categories are derived from the following source mappings:

\begin{itemize}
    \item \textbf{C1 (Task-Constraint Violation)} corresponds to category~1.1 Disobey Task Specification from the initial taxonomy.
    \item \textbf{C2 (Repeated or Stalled Progress)} merges categories~1.3 Step Repetition and~1.5 Unaware of Termination Conditions.
    \item \textbf{C3 (Unresolved Runtime Condition)} originates from categories~2.2 Fail to Ask for Clarification and~2.3 Task Derailment, and is extended in this work to cover execution or context failures.
    \item \textbf{C4 (Plan--Action--Outcome Inconsistency)} corresponds to category~2.6 Action--Reasoning Mismatch.
\end{itemize}

The boundary between C2 and C3 receives an additional decision rule because repetition is especially interpretation-sensitive. C2 requires repeated or semantically equivalent behavior after the trace has established that the current plan cannot succeed. A first failed action, one justified retry, a retry based on new parameters or evidence, or a substantive refinement is insufficient. An unresolved dependency without repetition is assigned C3 rather than C2.

\noindent{\bf Failure-node localization.}
Annotators choose the earliest trace-supported node at which an error is both actionable and connected through the recorded execution to the evaluator-confirmed failure. An \texttt{agent\_action} node is used when an agent decision commits the error; a \texttt{tool\_call} node when an erroneous or unjustified invocation commits it; a \texttt{tool\_result} node when a returned result or execution failure first becomes a decisive unhandled state; and a \texttt{final\_result} node when the error is introduced, or becomes decidable, only in the final response. Later repetitions or propagations of the same failure do not replace the earliest actionable node. Multiple symptoms at that node are encoded as multiple categories, not multiple primary nodes.

\noindent{\bf Annotation records and confidence.}
For each case, an annotator records the complete category set, primary category, failure-node ID and type, confidence, rationale, and supporting trace evidence. Confidence describes evidential clarity rather than a probability or voting weight: \textit{high} denotes direct support without a comparably supported alternative, \textit{medium} denotes a plausible alternative interpretation, and \textit{low} denotes insufficient evidence for a defensible tuple without further inspection. Confidence is retained as review metadata and does not mechanically affect adjudication.

\subsection{Annotation Procedure}

Four annotators with backgrounds in software engineering participated in the annotation process. Before annotating the full dataset, each annotator completed a calibration procedure consisting of reference-labeled examples and practice trajectories to establish consistent interpretation of the category definitions and localization rules.

Three annotators independently annotated each case. The fourth annotator then examined the complete traces, all three independent annotations, and the supporting evidence records, and produced the final annotation for each case. The final annotation consists of the strict tuple
\[
\begin{aligned}
(&\text{exact category set},\ \text{primary category},\\
 &\text{node ID},\ \text{node type}).
\end{aligned}
\]

Table~\ref{tab:adjudication-outcomes} summarizes the relationship between the initial independent annotations and the final annotation.

\begin{table}[t]
    \centering
    \small
    \caption{Relation between the three independent annotations and the final annotation. ``Proposed'' denotes a complete strict tuple submitted by at least one annotator.}
    \label{tab:adjudication-outcomes}
    \resizebox{\columnwidth}{!}{
    \begin{tabular}{lrl}
        \toprule
        Initial pattern & Cases & Outcome relative to final annotation \\
        \midrule
        Unanimous & 210 & all three retained 187; modified 23 \\
        2--1 split & 214 & majority retained 124; minority retained 48; new tuple 42 \\
        All distinct & 112 & one proposed tuple retained 77; new tuple 35 \\
        \bottomrule
    \end{tabular}}
\end{table}

The final annotation is supported in full by three initial annotators in 187 cases (34.89\%), two in 124 (23.13\%), one in 125 (23.32\%), and none in 100 (18.66\%). The final category set equals the three-annotator union in 423 cases (78.92\%) and differs in 113 (21.08\%). Across the latter decisions, the fourth annotator adds 123 case--category assignments absent from all initial annotations and removes 69 assignments proposed by at least one annotator, including 16 C1 assignments initially supported by all three. These bidirectional changes show that the final annotation re-evaluates trace evidence rather than merely aggregating votes.

\subsection{Inter-Annotator Reliability}

Reliability is computed from the three initial annotations for all 536 cases. Primary categories and node types are nominal variables. For the non-exclusive taxonomy, we calculate Fleiss' $\kappa$ both for the exact category combination as a nominal outcome and for the presence of each C1--C4 category as a binary decision. Percentile 95\% confidence intervals use 20,000 case-level bootstrap resamples with seed \texttt{20260722}; resampling cases preserves the three associated ratings. Final annotations are excluded from reliability calculations.

\begin{table}[H]
    \centering
    \small
    \caption{Agreement among the three independent human annotators. Pairwise agreement averages the three annotator pairs; 3/3 agreement is the number of cases in which all annotators make the same decision.}
    \label{tab:human-annotation-agreement}
    \setlength{\tabcolsep}{2.5pt}
    \renewcommand{\arraystretch}{1.12}
    \begin{tabularx}{\columnwidth}{
        @{}>{\raggedright\arraybackslash}X
        >{\centering\arraybackslash}p{0.27\columnwidth}
        >{\centering\arraybackslash}p{0.17\columnwidth}
        >{\centering\arraybackslash}p{0.20\columnwidth}@{}}
        \toprule
        Target &
        Fleiss' $\kappa$ (95\% CI) &
        Pairwise &
        Unanimous (3/3) \\
        \midrule
        Primary category
        & 0.622 (0.545--0.692)
        & 89.86\%
        & 459/536 (85.63\%) \\
        Exact multi-label category set
        & 0.574 (0.535--0.612)
        & 64.74\%
        & 280/536 (52.24\%) \\
        C1 presence
        & 0.696 (0.617--0.769)
        & 93.91\%
        & 487/536 (90.86\%) \\
        C2 presence
        & 0.374 (0.306--0.440)
        & 79.23\%
        & 369/536 (68.84\%) \\
        C3 presence
        & 0.718 (0.671--0.764)
        & 86.19\%
        & 425/536 (79.29\%) \\
        C4 presence
        & 0.688 (0.628--0.745)
        & 88.43\%
        & 443/536 (82.65\%) \\
        Failure-node type
        & 0.811 (0.759--0.857)
        & 93.97\%
        & 488/536 (91.04\%) \\
        \bottomrule
    \end{tabularx}
\end{table}

Failure-node IDs are case-specific and do not share a nominal category space, so Fleiss' $\kappa$ is not defined for this target. Exact agreement is reported instead: all three annotators select the same node in 396/536 cases (73.88\%; Wilson 95\% CI: 70.00\%--77.42\%), and at least two select the same node in 514/536 (95.90\%). For the complete strict tuple, all three annotations agree in 210/536 cases (39.18\%; Wilson 95\% CI: 35.14\%--43.37\%), and at least two agree in 424/536 (79.10\%).

C2 has the lowest category-specific reliability ($\kappa=0.374$) and accounts for 86 of the 123 added category decisions in the final annotation. This concentration identifies the repeated-or-stalled-progress boundary as the principal taxonomy-level ambiguity and motivates the explicit C2 decision rule above.

\subsection{Final Composition and Integrity Checks}

The 536 finalized cases contain 1,482 category assignments, with mean category cardinality 2.77. The primary-category distribution is C1 = 125 (23.32\%), C2 = 69 (12.87\%), C3 = 203 (37.87\%), and C4 = 139 (25.93\%). Multi-label marginal counts are C1 = 439, C2 = 288, C3 = 289, and C4 = 466; because categories are non-exclusive, these counts exceed the number of cases.

Every selected node is validated against its corresponding event graph. Validation requires the graph to exist, the node ID to occur in the graph, and the recorded node type to match the graph node. All 1,608 initial selections and 536 final selections satisfy these checks, for 2,144/2,144 validated selections in total.

\ds\ is scoped to two benchmark sources, three MASs, and 536 evaluator-confirmed failures with complete trace and graph artifacts. The finalized fields are human-adjudicated references, not outputs from the LLM annotation procedure evaluated later. We release the linked traces and graphs, all initial annotations and evidence records, and the adjudicated fields to support independent inspection; broader independent re-annotation remains future work.

\section{Execution and Statistical Conventions}
\label{app:experimental-setup}

\subsection{Repair Conditions and Execution Limits}

Paired methods are run on matched source failures. The task input, MAS implementation, recorded model alias, temperature, and native evaluator are held fixed within a pair. Conditions differ only in the information supplied by the repair prompt, the intervention location, and whether and where the recorded trajectory is replayed. Multiple internal model calls, tool calls, or technical retries within one top-level invocation are not treated as independent observations.

For RQ1, each source failure is evaluated through three executions per method. For RQ2, each task-level method receives up to three complete attempts and stops after its first evaluator-accepted result. For RQ3, each node-level method receives one selective-replay intervention, whereas the task-level baselines retain their up-to-three-attempt budget. A case is considered repaired if at least one permitted attempt passes the native evaluator. Actual API-call counts may differ because the methods generate different trajectories; efficiency is therefore measured rather than inferred from the nominal attempt budgets.

\subsection{Framework and Model Records}

The study uses AG2 0.13.3. The vendored lock installs the CrewAI main package as an editable project and therefore does not attach a version field to that lock entry; the main package's own \path{crewai/__init__.py} reports version 1.14.7a3, and its project metadata pins \texttt{crewai-core} and \texttt{crewai-cli} to the same version. The separately named \texttt{crewai-tools} package in the vendored workspace also reports version 1.14.7a3. The Magentic-One adapter uses AutoGen AgentChat 0.7.5. Its execution path imports \texttt{AssistantAgent} and \texttt{MagenticOneGroupChat} from \path{autogen_agentchat}. The model client, \texttt{OpenAIChatCompletionClient}, comes from \path{autogen_ext}. The adapter does not import \texttt{pyautogen}; the vendored lock records \texttt{pyautogen} 0.10.0 only as a meta-package whose dependency is \texttt{autogen-agentchat} 0.7.5.

The repair experiments were conducted from June 15 to June 16, 2026 UTC. Per-MAS execution windows are retained in the artifact.

All requests use temperature 0.00 through an OpenAI-compatible hosted endpoint. The requested model alias is \path{deepseek-v4-flash}. These values describe the recorded client request, not a verified immutable model snapshot. The experiment records do not contain a confirmed upstream provider identity or server-side revision. We therefore avoid attributing reproducibility to the alias or temperature setting alone.

\subsection{Statistical Procedures}

For a repair proportion $\hat{p}$, we report a 95\% Wilson confidence interval. For two methods evaluated on matched source failures, the effect size is the absolute repair-rate difference
\[
\Delta = 100(\hat{p}_{\mathrm s}-\hat{p}_{\mathrm b})
\]
in percentage points. Its 95\% confidence interval is estimated from 10,000 case-level paired bootstrap resamples using seed \texttt{20260701}. The two method outcomes for a case are resampled together to preserve pairing.

RQ1 compares \sys Replay with Unguided Full Rerun using a two-sided exact McNemar test on the paired per-case $\mathrm{rep}_3$ outcomes. RQ2 compares Self-Reflection and Critic-Agent separately with Unguided Full Rerun within each MAS; the two raw $p$-values form one prespecified Holm family per MAS. RQ3 compares Suspicious-Node Intervention with Unguided Full Rerun, Self-Reflection, Critic-Agent, Last-Node Intervention, and Random-Node Intervention within each MAS; the five raw $p$-values form one prespecified Holm family per MAS. Since there are three MASs, RQ3 contains three separate five-comparison Holm families totalling 15 tests. Holm-adjusted values are denoted by $p_{\mathrm H}$.

For every repair-method comparison, we report the matched denominator, both repair counts and rates, $\Delta$ and its paired confidence interval, and the applicable adjusted value. Each source failure contributes at most one binary outcome per method to a comparison; internal steps, calls, and technical retries are not analyzed as independent samples.

\subsection{Paired Comparison Results}
\label{app:paired-comparison-results}

All confidence intervals reported below are percentile intervals obtained from 10,000 case-level paired bootstrap resamples. McNemar tests are exact and two-sided. System and framework errors remain in the denominator and are counted as unsuccessful repairs. For RQ2, Holm correction is applied separately within each MAS to the two comparisons of Self-Reflection and Critic-Agent against Unguided Full Rerun. For RQ3, Holm correction is applied separately within each MAS to the five comparisons involving Suspicious-Node Intervention.

\begin{table*}[t]
\centering
\small
\caption{Paired RQ2 comparisons against Unguided Full Rerun. $\Delta$ is the repair-rate difference between the evaluated method and Unguided Full Rerun. $n_{10}/n_{01}$ denotes Method-only/Rerun-only repairs. Each Holm family contains the Self-Reflection and Critic-Agent comparisons within one MAS.}
\label{tab:rq2-paired-statistics}
\resizebox{\textwidth}{!}{
\begin{tabular}{llcccccc}
\toprule
MAS & Method & Method pass@3 & Rerun pass@3 &
$\Delta$ pp [95\% CI] & $n_{10}/n_{01}$ & Raw $p$ & $p_{\mathrm H}$ \\
\midrule
AG2
& Self-Reflection
& 8/171 (4.68\%)
& 14/171 (8.19\%)
& $-3.51$ [$-8.19$, 1.17]
& 5/11 & 0.210 & 0.237 \\

AG2
& Critic-Agent
& 7/171 (4.09\%)
& 14/171 (8.19\%)
& $-4.09$ [$-8.77$, 0.00]
& 4/11 & 0.118 & 0.237 \\

Magentic-One
& Self-Reflection
& 11/181 (6.08\%)
& 16/181 (8.84\%)
& $-2.76$ [$-7.73$, 2.21]
& 8/13 & 0.383 & 0.383 \\

Magentic-One
& Critic-Agent
& 8/181 (4.42\%)
& 16/181 (8.84\%)
& $-4.42$ [$-9.39$, 0.00]
& 6/14 & 0.115 & 0.231 \\

CrewAI
& Self-Reflection
& 4/184 (2.17\%)
& 7/184 (3.80\%)
& $-1.63$ [$-4.35$, 1.09]
& 2/5 & 0.453 & 0.906 \\

CrewAI
& Critic-Agent
& 5/184 (2.72\%)
& 7/184 (3.80\%)
& $-1.09$ [$-4.35$, 1.63]
& 3/5 & 0.727 & 0.906 \\
\bottomrule
\end{tabular}}
\end{table*}

Across all six RQ2 comparisons, the repair-rate differences relative to Unguided Full Rerun are negative, and none is statistically significant after within-MAS Holm correction. Thus, the results provide no evidence that Self-Reflection or Critic-Agent outperforms unguided rerunning under the same three-attempt budget.

\begin{table*}[t]
\centering
\small
\caption{Paired RQ3 comparisons. $\Delta$ is the repair-rate difference between Suspicious-Node Intervention and the comparator. $n_{10}/n_{01}$ denotes Suspicious-only/Comparator-only repairs. Suspicious-Node, Random-Node, and Last-Node use pass@1, whereas the three task-level comparators use pass@3, consistent with the budgets in the main paper. Each Holm family contains the five comparisons within one MAS.}
\label{tab:rq3-paired-statistics}
\resizebox{\textwidth}{!}{
\begin{tabular}{llcccccc}
\toprule
MAS & Comparator & Suspicious-Node & Comparator &
$\Delta$ pp [95\% CI] & $n_{10}/n_{01}$ & Raw $p$ & $p_{\mathrm H}$ \\
\midrule
AG2
& Unguided Full Rerun
& 28/171 (16.37\%)
& 14/171 (8.19\%)
& $+8.19$ [1.75, 14.62]
& 24/10 & 0.0243 & 0.0243 \\

AG2
& Self-Reflection
& 28/171 (16.37\%)
& 8/171 (4.68\%)
& $+11.70$ [5.26, 18.13]
& 26/6 & $5.35\times10^{-4}$ & $1.07\times10^{-3}$ \\

AG2
& Critic-Agent
& 28/171 (16.37\%)
& 7/171 (4.09\%)
& $+12.28$ [7.02, 18.13]
& 24/3 & $4.92\times10^{-5}$ & $1.48\times10^{-4}$ \\

AG2
& Random-Node
& 28/171 (16.37\%)
& 4/171 (2.34\%)
& $+14.04$ [8.19, 19.88]
& 27/3 & $8.43\times10^{-6}$ & $3.37\times10^{-5}$ \\

AG2
& Last-Node
& 28/171 (16.37\%)
& 1/171 (0.58\%)
& $+15.79$ [10.53, 21.64]
& 27/0 & $1.49\times10^{-8}$ & $7.45\times10^{-8}$ \\
\midrule
Magentic-One
& Unguided Full Rerun
& 34/181 (18.78\%)
& 16/181 (8.84\%)
& $+9.94$ [3.87, 16.02]
& 27/9 & 0.00393 & 0.00393 \\

Magentic-One
& Self-Reflection
& 34/181 (18.78\%)
& 11/181 (6.08\%)
& $+12.71$ [6.63, 18.78]
& 29/6 & $1.17\times10^{-4}$ & $2.34\times10^{-4}$ \\

Magentic-One
& Critic-Agent
& 34/181 (18.78\%)
& 8/181 (4.42\%)
& $+14.36$ [8.29, 20.44]
& 30/4 & $6.16\times10^{-6}$ & $1.85\times10^{-5}$ \\

Magentic-One
& Random-Node
& 34/181 (18.78\%)
& 7/181 (3.87\%)
& $+14.92$ [9.39, 20.44]
& 29/2 & $4.63\times10^{-7}$ & $1.85\times10^{-6}$ \\

Magentic-One
& Last-Node
& 34/181 (18.78\%)
& 3/181 (1.66\%)
& $+17.13$ [11.05, 23.20]
& 33/2 & $3.67\times10^{-8}$ & $1.84\times10^{-7}$ \\
\midrule
CrewAI
& Unguided Full Rerun
& 46/184 (25.00\%)
& 7/184 (3.80\%)
& $+21.20$ [14.67, 27.72]
& 43/4 & $2.78\times10^{-9}$ & $5.56\times10^{-9}$ \\

CrewAI
& Self-Reflection
& 46/184 (25.00\%)
& 4/184 (2.17\%)
& $+22.83$ [16.85, 29.35]
& 43/1 & $5.12\times10^{-12}$ & $2.05\times10^{-11}$ \\

CrewAI
& Critic-Agent
& 46/184 (25.00\%)
& 5/184 (2.72\%)
& $+22.28$ [15.76, 28.80]
& 43/2 & $5.89\times10^{-11}$ & $1.77\times10^{-10}$ \\

CrewAI
& Random-Node
& 46/184 (25.00\%)
& 9/184 (4.89\%)
& $+20.11$ [13.59, 26.63]
& 40/3 & $3.02\times10^{-9}$ & $5.56\times10^{-9}$ \\

CrewAI
& Last-Node
& 46/184 (25.00\%)
& 3/184 (1.63\%)
& $+23.37$ [17.39, 29.89]
& 44/1 & $2.61\times10^{-12}$ & $1.31\times10^{-11}$ \\
\bottomrule
\end{tabular}}
\end{table*}

Across all 15 RQ3 comparisons, Suspicious-Node Intervention achieves a positive repair-rate difference whose 95\% confidence interval excludes zero. All comparisons remain statistically significant after within-MAS Holm correction ($p_{\mathrm H}<0.05$). Thus, the advantage of Suspicious-Node Intervention is consistent across the three MAS architectures and against both task-level and node-level controls.

\subsection{API-Call Efficiency}
\label{app:api-call-efficiency}

Table~\ref{tab:api-call-efficiency} reports API-call efficiency as a secondary descriptive measure. A repair is counted only when \texttt{recovered=true}, indicating that the generated execution is accepted by the native evaluator. Cases ending in system or framework errors are counted as non-repairs, and any API calls made before those errors remain in the API-call total. For method $m$, we compute
\begin{equation}
\mathrm{RepairsPer1K}(m)
=
\frac{\mathrm{Repairs}(m)}
     {\mathrm{APICalls}(m)}
\times 1000,
\end{equation}
and
\begin{equation}
\mathrm{CallsPerRepair}(m)
=
\frac{\mathrm{APICalls}(m)}
     {\mathrm{Repairs}(m)}.
\end{equation}
Because methods may generate trajectories of different lengths and receive different attempt budgets, these measures are compared primarily within each MAS and are not treated as compute-matched causal estimates.

\begin{table*}[t]
\centering
\footnotesize
\caption{API-call efficiency on the three MASs evaluated in the main paper. Repairs count native-evaluator-accepted executions. Higher Repairs/1K calls and lower Calls/repair indicate greater repair yield per API call. Bold marks the best value within each MAS.}
\label{tab:api-call-efficiency}
\setlength{\tabcolsep}{9pt}
\renewcommand{\arraystretch}{0.96}
\begin{tabular}{@{}llrrrr@{}}
\toprule
MAS & Method & API calls & Repairs &
Repairs/1K calls & Calls/repair \\
\midrule
AG2
& Last-Node
& 249 & 1 & 4.0 & 249.0 \\
AG2
& Random-Node
& 293 & 4 & 13.7 & 73.3 \\
AG2
& Critic-Agent
& 295 & 7 & 23.7 & 42.1 \\
AG2
& Self-Reflection
& 314 & 8 & 25.5 & 39.3 \\
AG2
& Unguided Full Rerun
& 400 & 14 & 35.0 & 28.6 \\
AG2
& Suspicious-Node
& 419 & 28 & \textbf{66.8} & \textbf{15.0} \\
\midrule
CrewAI
& Last-Node
& 227 & 3 & 13.2 & 75.7 \\
CrewAI
& Random-Node
& 251 & 9 & 35.9 & 27.9 \\
CrewAI
& Critic-Agent
& 256 & 5 & 19.5 & 51.2 \\
CrewAI
& Self-Reflection
& 258 & 4 & 15.5 & 64.5 \\
CrewAI
& Unguided Full Rerun
& 254 & 7 & 27.6 & 36.3 \\
CrewAI
& Suspicious-Node
& 466 & 46 & \textbf{98.7} & \textbf{10.1} \\
\midrule
Magentic-One
& Last-Node
& 239 & 3 & 12.6 & 79.7 \\
Magentic-One
& Random-Node
& 278 & 7 & 25.2 & 39.7 \\
Magentic-One
& Critic-Agent
& 324 & 8 & 24.7 & 40.5 \\
Magentic-One
& Self-Reflection
& 343 & 11 & 32.1 & 31.2 \\
Magentic-One
& Unguided Full Rerun
& 367 & 16 & 43.6 & 22.9 \\
Magentic-One
& Suspicious-Node
& 443 & 34 & \textbf{76.7} & \textbf{13.0} \\
\bottomrule
\end{tabular}
\end{table*}

Across AG2, CrewAI, and Magentic-One, Suspicious-Node achieves both the highest number of accepted repairs per 1,000 API calls and the lowest number of API calls per repair. These results provide consistent secondary evidence that symptom-guided localized intervention yields greater repair efficiency than the task-level and node-level controls within each MAS.

\subsection{Implementation Environment}

The experiments are orchestrated with Python 3.11.15 on Windows 11 using an Intel Core i9-14900HX processor, 32~GB of RAM, and an NVIDIA GeForce RTX 4060 Laptop GPU with 8~GB of VRAM. Model inference is performed by the hosted endpoint. The local workstation runs the MAS frameworks and tools, trace capture, \sys replay, native evaluation, and statistical analysis.

\section{Reproduction Settings and Fidelity}
\label{app:rq1-details}

\subsection{Mechanism-Level Failure Equivalence}

The formal equivalence test used by the method-blinded judges is retained below. For attempt $a$ of method $m$, the category-matching events are
\begin{equation}
\mathcal{M}^{c}_{i,a,m}
=
\left\{\widehat e\in\widehat\tau_{i,a}^{m}:
\widehat c(\widehat e)=c_i\right\}.
\end{equation}
The mechanism-matching and role-matching sets are
\begin{align}
\mathcal{M}^{m}_{i,a,m}
&=
\left\{\widehat e\in\widehat\tau_{i,a}^{m}:
\widehat m(\widehat e)\equiv m_i\right\},\\
\mathcal{M}^{\rho}_{i,a,m}
&=
\left\{\widehat e\in\widehat\tau_{i,a}^{m}:
\widehat\rho(\widehat e)\sim\rho_i\right\}.
\end{align}
Their intersection contains events satisfying all three semantic criteria:
\begin{equation}
\mathcal{M}_{i,a,m}
=
\mathcal{M}^{c}_{i,a,m}
\cap\mathcal{M}^{m}_{i,a,m}
\cap\mathcal{M}^{\rho}_{i,a,m}.
\end{equation}
The attempt reproduces the source mechanism only when at least one such event also has grounded evidence:
\begin{equation}
z_{i,a}^{m}
=
\mathbf{1}\!\left[
\exists\widehat e\in\mathcal{M}_{i,a,m}:
\operatorname{Grounded}(\widehat E_{\widehat e},E_i)
\right].
\end{equation}
Here, $c_i$, $m_i$, $\rho_i$, and $E_i$ denote the source category, case-specific mechanism, semantic execution role, and supporting evidence. The relations $\equiv$ and $\sim$ express mechanism equivalence and semantic-role correspondence rather than raw node equality.

\subsection{Reproduction Metrics}

Let $N$ be the number of source cases. Per-execution reproduction is
\begin{equation}
\mathrm{rep}_1(m)
=
\frac{1}{3N}
\sum_{i=1}^{N}\sum_{a=1}^{3}z_{i,a}^{m}.
\end{equation}
Reproduction in all three attempts is
\begin{equation}
\mathrm{rep}_3(m)
=
\frac{1}{N}
\sum_{i=1}^{N}
\prod_{a=1}^{3}z_{i,a}^{m}.
\end{equation}

\subsection{Replay-Fidelity Audit}

Failure recurrence does not itself establish that replay preserved the historical context. Before the target, every model or tool request is matched against the ordered replay bundle using its event position and canonicalized request content. A missing, out-of-order, or non-matching request terminates replay rather than silently continuing.

We separately compare the materialized replay graph with the source prefix. Across 536 cases and three attempts per case, every replay plan uses the exact target ID from the corresponding human-adjudicated source annotation (1,608/1,608). Every attempt also preserves the parent/edge topology of the reused prefix, and all 6,159 materialized reused nodes match their source content hashes (6,159/6,159). Attempt-level prefix exactness is therefore 1,608/1,608. Because the target is supplied from the human-adjudicated annotation, this audit measures replay fidelity rather than automatic localization accuracy.

These audits rule out modified recorded-prefix content as the source of reproduction below 100.00\%. Remaining variation may arise at the regenerated target, in subsequent live execution, or from external state outside the replay boundary.

\section{Task-Level Repair Aggregation}
\label{app:rq2-details}

The main paper defines the task-level strategies, evaluation sets, budgets, and results. We retain only the formal $\mathrm{pass@3}$ aggregation. Let $\mathcal{F}$ denote an evaluation set of initially failed executions, $y_{i,a}^{m}$ the evaluator outcome of attempt $a$, and $A_i^{m}\leq3$ the attempts executed before success or exhaustion. The terminal case outcome is
\begin{equation}
R_i^{m}
=
\mathbf{1}\left[
\max_{1\leq a\leq A_i^{m}}y_{i,a}^{m}=1
\right].
\end{equation}
Cumulative repair is
\begin{equation}
\mathrm{pass@3}(m)
=
\frac{1}{|\mathcal{F}|}
\sum_{i\in\mathcal{F}}R_i^{m}.
\end{equation}

\subsection{Baseline Input Boundaries and Prompt Templates}
\label{app:rq2-prompts}

The three task-level baselines are evaluated on the same 536 source failures from AG2, CrewAI, and Magentic-One. All three receive the task specification, task identifier, benchmark and split identifiers, and the available start URLs or tools. Self-Reflection and Critic-Agent additionally receive the final answer from the original failed execution. None of the baselines receives the source trajectory or event graph, failure location, C1--C4 category, trace evidence, reference or gold answer, or evaluator information.

For every recovery attempt, Self-Reflection and Critic-Agent reuse the original failed execution's final answer rather than the output of a preceding recovery attempt. Critic-Agent implements the dedicated critic role described in the main paper through a fixed critic-role prompt within each complete rerun attempt; it does not introduce an additional standalone critic-model call. Apart from their method-specific feedback, all three baselines use the same task information, output suffix, attempt budget, model configuration, and native evaluator.

\noindent{\bf Unguided Full Rerun.}
\begin{lstlisting}[style=MASDTrace,numbers=none]
Mode: fail-then-rerun
Task id: {task_id}
Benchmark: {source_benchmark} / {source_split}
Task: {task}
Start URLs or tools: {start_urls}
This is an unguided rerun after an initial failed or
uncertain attempt.
Solve from scratch.
{output_suffix}
\end{lstlisting}

\noindent{\bf Self-Reflection.}
\begin{lstlisting}[style=MASDTrace,numbers=none]
Mode: self-reflection
Task id: {task_id}
Benchmark: {source_benchmark} / {source_split}
Task: {task}
Start URLs or tools: {start_urls}
Previous final answer:
{previous_final_answer}
Reflect on why the previous attempt may be incomplete
or wrong, then solve the task again.
{output_suffix}
\end{lstlisting}

\noindent{\bf Critic-Agent.}
\begin{lstlisting}[style=MASDTrace,numbers=none]
Mode: critic-agent
Task id: {task_id}
Benchmark: {source_benchmark} / {source_split}
Task: {task}
Start URLs or tools: {start_urls}
Previous final answer:
{previous_final_answer}
Critic feedback: verify assumptions, check whether the
cited evidence is sufficient, and correct any unsupported
or stale claims before producing the new answer.
{output_suffix}
\end{lstlisting}

For the WebArena-Verified and AssistantBench tasks used in this study, \texttt{\{output\_suffix\}} is instantiated as follows:

\begin{lstlisting}[style=MASDTrace,numbers=none]
Return only the final answer and a short note about what
evidence would be needed.
\end{lstlisting}

Because the three task-level baselines share the same task inputs, output suffix, model configuration, evaluator, and three-attempt budget, their comparison isolates whether access to the original failed answer and generic reflection or critic guidance provides an advantage over unguided rerunning. The paired results in Appendix~\ref{app:paired-comparison-results} show no such advantage for Self-Reflection or Critic-Agent.

\section{Detection Rules, Anchor Ranking, and Intervention Prompts}
\label{app:rq3-details}

\begin{figure*}[t]
    \centering
    \paperfigure[
        width=\textwidth,
        keepaspectratio
    ]{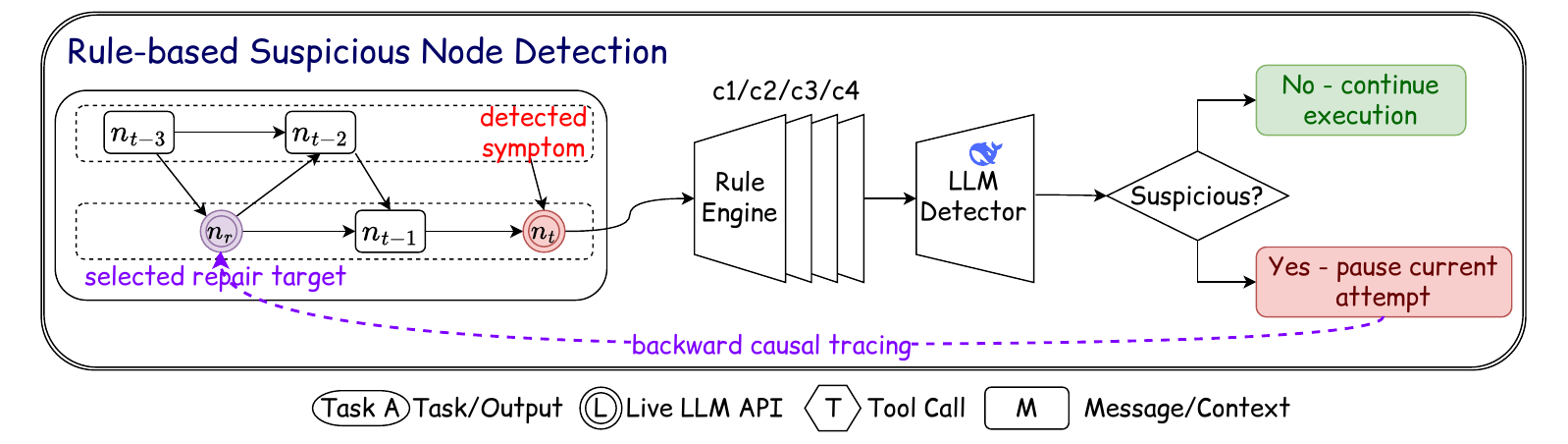}
    \caption{Suspicious-Node Intervention pipeline. Deterministic rules and a semantic judge jointly evaluate each newly completed node for the C1--C4 symptoms. When the fused suspicion score exceeds the predefined threshold, the ongoing MAS execution is suspended. \sys then reconstructs the execution prefix preceding the selected intervention anchor, injects symptom-conditioned repair guidance derived from the triggered rules, and resumes live execution to generate a new downstream trajectory.}
    \label{fig:suspicious-node}
    \Description{The suspicious-node intervention pipeline, including rule-based symptom detection, semantic scoring, anchor selection, replay, and live repair.}
\end{figure*}

\subsection{Rule-First Symptom Detection}

Let $G=(V,E)$ be the event-dependency graph, $T$ the task specification, and $\tau_{<v}$ the trace preceding node $v$. For category $C_k$, define its trigger indicator as
\begin{equation}
b_k(v)
=
\bigvee_{r\in\mathcal{R}_k}
r(v,\tau_{<v},T).
\end{equation}
The deterministic candidate set collects the triggered categories:
\begin{equation}
L_v^{\mathrm r}
=
\left\{C_k:b_k(v)=1\right\}.
\end{equation}

\begin{table*}[t]
\caption{Deterministic conditions used to generate trace-localized symptom candidates. The semantic judge may reject a candidate but cannot introduce a category outside this set.}
\label{tab:rq3-rules}
\centering
\footnotesize
\setlength{\tabcolsep}{4pt}
\begin{tabular}{>{\raggedright\arraybackslash}p{0.16\textwidth}
                >{\raggedright\arraybackslash}p{0.22\textwidth}
                >{\raggedright\arraybackslash}p{0.52\textwidth}}
\toprule
Category & Operational meaning & Deterministic candidate conditions \\
\midrule
C1: Task-constraint violation &
The behavior conflicts with an explicit task constraint or output requirement. &
The node violates a stated constraint, produces an incompatible output format, contains a placeholder or skipped action, or returns an unresolved request instead of the required result. \\

C2: Repeated or stalled progress &
The history repeats while the objective remains unfinished. &
The normalized action fingerprint or tool URL matches a preceding node; the node retries without new parameters or evidence; or another final-style output follows without an action that advances the unfinished objective. \\

C3: Unresolved runtime condition &
The node contains or continues from an unresolved execution, access, or upstream condition. &
The node records an error, timeout, or failed tool result; reports inaccessible required evidence; or proceeds toward a final answer while an error remains in the preceding trace. \\

C4: Plan--action--outcome inconsistency &
An explicit plan or claim conflicts with the executed action or observed outcome. &
A browser action lacks a URL or uses an invalid \texttt{view-source:} URL; structured data is requested but HTML is returned; completion follows an unsuccessful action; or a tool-dependent task is answered without successful tool evidence. \\
\bottomrule
\end{tabular}
\end{table*}

Algorithm~\ref{alg:symptom-detection} applies these conditions to every completed node. Nodes without a deterministic trigger receive zero suspiciousness without invoking the semantic judge. For triggered nodes, the judge may confirm or reject supplied categories but must satisfy
\begin{equation}
L_v^{\mathrm f}\subseteq L_v^{\mathrm r}.
\end{equation}

\begin{algorithm}[t]
\caption{Rule-Gated Symptom Detection}
\label{alg:symptom-detection}
\small
\begin{algorithmic}[1]
\REQUIRE Task $T$, event-dependency graph $G$, observed order $O$
\REQUIRE Rule sets $\{\mathcal{R}_1,\ldots,\mathcal{R}_4\}$ and semantic judge $\mathcal{J}$
\ENSURE Confirmed candidate set $\mathcal{C}$
\STATE $\mathcal{C}\leftarrow\emptyset$
\FOR{each completed node $v$ in $O$}
    \STATE Construct the diagnostic view from $(T,v,\tau_{<v},G)$
    \STATE $L_v^{\mathrm r}\leftarrow\emptyset$; $E_v^{\mathrm r}\leftarrow\emptyset$
    \FOR{$k=1$ to $4$}
        \FOR{each rule $r\in\mathcal{R}_k$}
            \IF{$r(v,\tau_{<v},T)=1$}
                \STATE Add $C_k$ and its matched evidence to $(L_v^{\mathrm r},E_v^{\mathrm r})$
            \ENDIF
        \ENDFOR
    \ENDFOR
    \IF{$L_v^{\mathrm r}\neq\emptyset$}
        \STATE Remove all reference-answer and evaluator information
        \STATE Query $\mathcal{J}$ for $(L_v^{\mathrm f},\widetilde c_{\mathrm{LLM}}(v),\xi_v)$
        \STATE $L_v^{\mathrm f}\leftarrow L_v^{\mathrm f}\cap L_v^{\mathrm r}$
        \IF{$L_v^{\mathrm f}\neq\emptyset$}
            \STATE $E_v^{\mathrm f}\leftarrow$ evidence in $E_v^{\mathrm r}$ supporting $L_v^{\mathrm f}$
            \STATE Add $(v,L_v^{\mathrm f},E_v^{\mathrm f},\widetilde c_{\mathrm{LLM}}(v))$ to $\mathcal{C}$
            \STATE Retain $\xi_v$ for audit only
        \ENDIF
    \ENDIF
\ENDFOR
\STATE \textbf{return} $\mathcal{C}$
\end{algorithmic}
\end{algorithm}

We write $L_v=L_v^{\mathrm f}$ for the confirmed category set used by the ranking procedure. Reference answers, evaluator verdicts, and evaluator-derived evidence are removed before verification, so both rule evidence and semantic judgments use only the task and runtime trace.

C1--C4 are non-exclusive operational symptoms. A category can denote a local error, unresolved dependency, or downstream manifestation, so it raises intervention priority without asserting that the node is the unique root cause.

\subsection{Repairability-Aware Anchor Selection}

A strongly suspicious node need not be the best intervention anchor: terminal nodes leave little suffix to repair, while propagation-only nodes may merely expose earlier failures. Let $c_{\mathrm l}(v)$ be the strongest locally supported confidence, $c_{\mathrm r}(v)$ include propagated evidence, $I_{\mathrm p}(v)$ indicate propagation-only evidence, and $\rho(v)\in[0,1]$ be normalized graph position.

The repairability score begins with node type and local evidence:
\begin{equation}
B_{\mathrm r}(v)
=
\beta_0+
\eta(v)+
\beta_{\mathrm l}c_{\mathrm l}(v).
\end{equation}
Symptom multiplicity contributes
\begin{equation}
G_{\mathrm{multi}}(v)
=
\beta_{\mathrm m}\mathbf{1}[|L_v|>1].
\end{equation}
The C2 and C4 category boost is
\begin{equation}
G_{\mathrm{cat}}(v)
=
\beta_{C_4}\mathbf{1}[C_4\in L_v]
+\beta_{C_2}\mathbf{1}[C_2\in L_v].
\end{equation}
Their sum forms the positive repairability adjustment:
\begin{equation}
G_{\mathrm r}(v)
=
G_{\mathrm{multi}}(v)+G_{\mathrm{cat}}(v).
\end{equation}
Propagation-only evidence and late graph position contribute the penalty
\begin{equation}
P_{\mathrm r}(v)
=
\beta_{\mathrm p}I_{\mathrm p}(v)
+\beta_{\pi}\rho(v).
\end{equation}
The complete repairability score is
\begin{equation}
S_{\mathrm r}(v)
=
\operatorname{clip}_{[0,1]}
\left(B_{\mathrm r}(v)+G_{\mathrm r}(v)-P_{\mathrm r}(v)\right).
\label{eq:repairability-score}
\end{equation}
Here, $\eta(v)$ is a node-type prior. Regenerable agent actions receive the highest prior, followed by tool calls and results, while final-result nodes receive the lowest.

Among replayable nodes, the selection score first combines repairability with local and propagated confidence:
\begin{equation}
B_{\mathrm s}(v)
=
\alpha_{\mathrm r}S_{\mathrm r}(v)
+\alpha_{\mathrm l}c_{\mathrm l}(v)
+\alpha_{\mathrm u}c_{\mathrm r}(v).
\end{equation}
The positive selection adjustment is
\begin{equation}
G_{\mathrm s}(v)
=
\alpha_{\mathrm n}\min(2,|L_v|)
+\alpha_{\mathrm a}I_{\mathrm a}(v),
\end{equation}
and the selection penalty is
\begin{equation}
P_{\mathrm s}(v)
=
\alpha_{\mathrm p}I_{\mathrm p}(v)
+\alpha_{\mathrm f}I_{\mathrm f}(v)
+\alpha_{\pi}\rho(v).
\end{equation}
The complete selection score is
\begin{equation}
S_{\mathrm s}(v)
=
\operatorname{clip}_{[0,1]}
\left(B_{\mathrm s}(v)+G_{\mathrm s}(v)-P_{\mathrm s}(v)\right).
\label{eq:selection-score}
\end{equation}
Here, $I_{\mathrm a}(v)$ indicates an actionable \texttt{agent\_action} or \texttt{tool\_call} boundary and $I_{\mathrm f}(v)$ indicates a \texttt{final\_result} node. The coefficient design prioritizes repairability, followed by local evidence and then propagated evidence.

Semantic confidence and the structured score are first mixed as
\begin{equation}
c_{\mathrm{mix}}(v)
=
0.55\,\widetilde{c}_{\mathrm{LLM}}(v)
+0.45\,S_{\mathrm s}(v).
\end{equation}
Multiple confirmed symptoms provide the bonus
\begin{equation}
b_{\mathrm{multi}}(v)
=
0.04\,\mathbf{1}[|L_v|>1].
\end{equation}
The fused confidence is therefore
\begin{equation}
c_{\mathrm f}(v)
=
\operatorname{clip}_{[0,1]}
\left(c_{\mathrm{mix}}(v)+b_{\mathrm{multi}}(v)\right).
\label{eq:fused-confidence}
\end{equation}
A replayable node is eligible when it has a confirmed category and
\begin{equation}
c_{\mathrm f}(v)\geq\theta,\qquad\theta=0.50.
\end{equation}
Algorithm~\ref{alg:targeted-intervention} implements the online intervention flow: nodes are processed in observed order, and the first eligible node whose fused score meets the threshold triggers replay and live repair. These quantities are ordinal evidence scores, not calibrated probabilities. Coefficients encode modeling priorities and were not optimized using validation repair outcomes; $\theta=0.50$ is a coverage-oriented gate rather than a globally optimal threshold.

\begin{algorithm}[t]
\caption{Online Suspicious-Node Intervention}
\label{alg:targeted-intervention}
\small
\begin{algorithmic}[1]
\REQUIRE Replay bundle $\mathcal{S}$, task $T$, candidate set $\mathcal{C}$, threshold $\theta$
\ENSURE Regenerated trace and native-evaluator outcome, or no intervention
\FOR{each $(v,L_v,E_v,\widetilde c_{\mathrm{LLM}}(v))\in\mathcal{C}$ in observed order}
    \STATE Derive local evidence, propagation, node-type, and position features
    \STATE Compute $S_{\mathrm r}(v)$ using Equation~\ref{eq:repairability-score}
    \STATE Compute $S_{\mathrm s}(v)$ using Equation~\ref{eq:selection-score}
    \STATE Compute $c_{\mathrm f}(v)$ using Equation~\ref{eq:fused-confidence}
    \IF{$v$ is a replayable LLM boundary and $c_{\mathrm f}(v)\geq\theta$}
        \STATE Build $\Delta^\star$ from the target ID, $L_v$, and rule-derived $E_v$
        \STATE Exclude the judge rationale and all evaluator-derived information
        \STATE $\widehat\tau\leftarrow\textsc{Replay}(\mathcal{S},\mathrm{selective},v,\Delta^\star)$ via Algorithm~\ref{alg:replay}
        \STATE $y\leftarrow\textsc{NativeEvaluator}(\widehat\tau)$
        \STATE \textbf{return} $(\widehat\tau,y)$
    \ENDIF
\ENDFOR
\STATE \textbf{return} no intervention
\end{algorithmic}
\end{algorithm}

\subsection{Intervention Prompts}

\noindent{\bf Suspicious-Node Intervention.}
The controller instantiates the following template: \textit{Reconsider the decision at target [target identifier]. Confirmed runtime symptoms: [categories]. Trace evidence: [rule-derived evidence]. Revise this step so that the remaining execution addresses the evidence and the task requirements, and regenerate any affected downstream actions or artifacts.} The free-form semantic-judge rationale, reference answer, evaluator verdict, and evaluator-derived evidence are not included.

\noindent{\bf Random-Node and Last-Node Intervention.}
Both controls receive the same generic template: \textit{Re-check the task requirements and the selected step. Correct any issue at this step, then regenerate the affected downstream actions or artifacts.} The template contains no symptom category or trace evidence.

\subsection{Metrics and Statistical Procedure}

Let $R_i^{(m)}$ indicate whether method $m$ repairs failed case $i$. For the single-intervention node-level methods,
\begin{equation}
\mathrm{pass@1}(m)
=
\frac{1}{N}
\sum_{i=1}^{N}R_i^{(m)}.
\end{equation}
Infrastructure and framework errors remain non-repairs; the case is the inferential unit. Statistical comparisons follow Appendix~\ref{app:experimental-setup}.

\subsection{Threshold Interpretation}

Selected-node scores in the primary 536-case evaluation set have unrounded minimum, 25th-percentile, median, 75th-percentile, and maximum values of 0.5003744, 0.6900606, 0.8910112, 0.9360944, and 1.0000000, respectively; these round to 0.50, 0.69, 0.89, 0.94, and 1.00. In a post-hoc threshold-as-abstention analysis that keeps each historical rank-0 target and its observed repair outcome fixed, $\theta=0.50$ retains 536/536 cases (100.00\% coverage), including 108 successful repairs, for 108/536 = 20.15\% conditional repair. At $\theta=0.95$, 78/536 cases are retained (14.55\% coverage), including 19 successful repairs, for 19/78 = 24.36\% conditional repair. This analysis neither reranks candidate nodes nor reruns repairs at alternative targets.

We interpret 0.50 as a minimum evidence gate. The configuration is practically effective but neither the threshold nor individual coefficients are claimed to be globally optimal. Systematic optimization and component-level ablation remain limitations.

\noindent\textit{Note on scoring parameters.}
The implemented selection coefficients, in the order used above, are
\begin{equation}
\begin{aligned}
(&\alpha_{\mathrm r},\alpha_{\mathrm l},\alpha_{\mathrm u},\alpha_{\mathrm n},
\alpha_{\mathrm a},\alpha_{\mathrm p},\alpha_{\mathrm f},\alpha_{\pi})\\
&=(0.48,0.32,0.12,0.04,0.08,0.28,0.18,0.12).
\end{aligned}
\end{equation}
Here, $\alpha_{\mathrm u}=0.12$ multiplies the raw confidence $c_{\mathrm r}(v)$, which may include propagated evidence, and $\alpha_{\mathrm n}=0.04$ is multiplied by $\min(2,|L_v|)$, so its contribution is capped at 0.08. The repairability coefficients are
\begin{equation}
\begin{aligned}
(&\beta_0,\beta_{\mathrm l},\beta_{\mathrm m},\beta_{C_2},
\beta_{C_4},\beta_{\mathrm p},\beta_{\pi})\\
&=(0.20,0.26,0.08,0.05,0.07,0.30,0.18),
\end{aligned}
\end{equation}
with the node-type prior
\begin{equation}
\eta(v)=
\begin{cases}
0.42, & \operatorname{type}(v)=\texttt{agent\_action},\\
0.36, & \operatorname{type}(v)=\texttt{tool\_call},\\
0.24, & \operatorname{type}(v)=\texttt{tool\_result},\\
-0.18, & \operatorname{type}(v)=\texttt{final\_result},\\
0, & \text{otherwise}.
\end{cases}
\end{equation}
The penalties appear with positive coefficients in $P_{\mathrm r}$ and $P_{\mathrm s}$ and are subtracted by Equations~\ref{eq:repairability-score} and~\ref{eq:selection-score}. The fusion weights 0.55 for semantic-judge confidence and 0.45 for the structured selection score are exact, as is the separate $+0.04$ bonus for more than one confirmed symptom. These values are hard-coded in \path{code/src/run_rq3_node_llm_judge.py} and enumerated in \path{if/data/rq3_node_judge_hyperparameters/tables/hyperparameter-inventory.csv}. They are manually specified heuristic design weights; no saved validation-set optimization or alternative-weight sweep exists, so we do not describe them as fitted, calibrated, or empirically optimal.

\subsection{Semantic Boundary of Suspiciousness}

To formalize the semantic-boundary audit summarized in the main paper, define
\begin{align}
\mathcal{E}_{\mathrm f}=\{&
\texttt{failed},\texttt{failure},\texttt{error},\notag\\
&\texttt{timeout},\texttt{timed\_out}\},
\end{align}
The node-level error indicator is
\begin{equation}
S_{\mathrm{node}}(v)
=
\mathbf{1}[\texttt{v.error}\neq\emptyset].
\end{equation}
The tool-result error indicator is
\begin{equation}
S_{\mathrm{tool}}(v)
=
\mathbf{1}[\texttt{v.tool\_result.error}\neq\emptyset],
\end{equation}
and the tool-status indicator is
\begin{equation}
S_{\mathrm{status}}(v)
=
\mathbf{1}[\texttt{v.tool\_result.status}\in\mathcal{E}_{\mathrm f}].
\end{equation}
The final runtime-failure signal combines the three indicators:
\begin{equation}
S(v)
=
\max\left\{S_{\mathrm{node}}(v),S_{\mathrm{tool}}(v),S_{\mathrm{status}}(v)\right\}.
\end{equation}


\section{Traceable Gold Failure Examples}
\label{app:gold-failure-examples}

\begin{tcolorbox}[
  enhanced,
  breakable,
  colback=MASDNavy!3!white,
  colframe=MASDNavy,
  boxrule=0.8pt,
  borderline north={2pt}{0pt}{MASDNavy},
  arc=1.5mm,
  title={\textbf{Corpus anchor and selection protocol}},
  fonttitle=\sffamily\bfseries,
  colbacktitle=MASDNavy,
  coltitle=white,
  before upper=\raggedright
]
The gold ledger contains \textbf{536 finalized, human-adjudicated failure
cases}: 462 from WebArena-Verified and 74 from AssistantBench. The corresponding
multi-agent-system distribution is 171 AG2, 184 CrewAI, and 181
Magentic-One cases.

We select one \textit{trace-verified} example from every observed primary
symptom category:
\masdpill{MASDBlue}{C1: 125/536}
\masdpill{MASDAmber}{C2: 69/536}
\masdpill{MASDRed}{C3: 203/536}
\masdpill{MASDPurple}{C4: 139/536}.
Thus, the examples cover all four symptom categories observed in the
primary distribution. Selection is stratified by category rather
than by favorable outcome.

Each example reports the immutable case, annotation, task, and node
identifiers; a field-preserving excerpt of the recorded trace; the
evidence-supported localization; and the relative path of the complete
raw-log bundle. Ellipses inside excerpts remove only payload that is irrelevant
to the stated attribution; the raw bundle remains the authoritative
record.
\end{tcolorbox}

\begin{table*}[t]
\centering
\small
\setlength{\tabcolsep}{6pt}
\caption{
Four stratified gold examples. ``Prevalence'' is computed from the
536-case finalized ledger by primary category. The displayed examples are qualitative
trace evidence, not estimates of model accuracy.
}
\label{tab:gold-example-overview}
\begin{tabular}{@{}lllllr@{}}
\toprule
\textbf{Stratum}
& \textbf{Selection role}
& \textbf{Case ID}
& \textbf{Benchmark / MAS}
& \textbf{Gold node kind}
& \textbf{Prevalence} \\
\midrule
\masdpill{MASDBlue}{C1}
& Standard clear positive
& \texttt{40fe6e29c9209dfd}
& AssistantBench / AG2
& \path{final_result}
& 23.32\% \\

\masdpill{MASDAmber}{C2}
& Complex multi-hop context
& \texttt{1aaa0ae52cbcf325}
& WebArena / AG2
& \path{tool_result}
& 12.87\% \\

\masdpill{MASDRed}{C3}
& Operational low-signal noise
& \texttt{06845b3bc9dad059}
& WebArena / AG2
& \path{tool_result}
& 37.87\% \\

\masdpill{MASDPurple}{C4}
& Plan--action inconsistency
& \texttt{11772545c348fb2a}
& WebArena / CrewAI
& \path{agent_action}
& 25.93\% \\
\bottomrule
\end{tabular}
\end{table*}


\begin{masdgold}
  {GOLD--C1}
  {Standard Clear Positive}
  {MASDBlue}

\masdmeta
  {40fe6e29c9209dfd}
  {9f6c08b2c4fe9155}
  {assistantbench_dev_9fce12bb25f697d6}
  {AssistantBench development split / AG2}
  {15 nodes; 3 recorded API interactions}
  {assistantbench_dev_9fce12bb25f697d6:attempt_000:snapshot:31ff9ace4b7fbae1:final_result}
  {data/initial_failure_logs/ag2/assistantbench_dev_9fce12bb25f697d6/attempt_000}

\begin{masdpanel}[colframe=MASDBlue!55]{Task prompt}
\textit{Which supermarkets within 2 blocks of Lincoln Park in Chicago
have ready-to-eat salad for under \$15?}
\end{masdpanel}

\begin{masdpanel}{Recorded execution path}
\begin{center}
\resizebox{\linewidth}{!}{%
\begin{tikzpicture}[node distance=3.1mm]
  \node[masdstep] (task)
    {Constrained\\entity query};
  \node[masdok, right=of task] (r1)
    {\path{tool_result_0001}\\HTTP 200};
  \node[masdok, right=of r1] (r2)
    {\path{tool_result_0002}\\Potash site\\HTTP 200};
  \node[masdfail, right=of r2] (final)
    {\path{final_result}\\another browse\\action};

  \draw[masdarrow] (task) -- (r1);
  \draw[masdarrow] (r1) -- (r2);
  \draw[masdarrow] (r2) -- (final);

  \node[
    below=1.2mm of final,
    text=MASDBlue,
    font=\sffamily\bfseries\scriptsize
  ] {GOLD FAILURE NODE};
\end{tikzpicture}%
}
\end{center}
\end{masdpanel}

\begin{masdpanel}{Field-preserving raw trace excerpt}
\begin{lstlisting}[style=MASDTrace]
case_id = "40fe6e29c9209dfd"
annotation_id = "9f6c08b2c4fe9155"

failure_node_id =
"assistantbench_dev_9fce12bb25f697d6:attempt_000:
 snapshot:31ff9ace4b7fbae1:final_result"

tool_result_0001.status = "success"
tool_result_0001.http_status = 200

tool_result_0002.status = "success"
tool_result_0002.http_status = 200
tool_result_0002.url =
"https://www.potashmarkets.com/"

final_result =
{"action":"browse",
 "url":"https://www.potashmarkets.com/",
 "reason":"Find Potash Markets locations and addresses
           near Lincoln Park."}

evaluator.reference_items =
["Potash Markets - Clark Street"]

evaluator.missing_reference_items =
["Potash Markets - Clark Street"]
\end{lstlisting}
\end{masdpanel}

\begin{masdverdict}{MASDBlue}{Attribution}
\masdpill{MASDBlue}{C1}
\textbf{Task-constraint violation / incomplete resolution.}

\noindent
\textbf{Erroneous node.}
The terminal \path{final_result} node is the first node at which the
trajectory irreversibly fails to satisfy the requested output contract.

\noindent
\textbf{Error committed.}
Instead of returning the name of a qualifying supermarket, the node
returns another browsing action. The trajectory therefore terminates
before converting successfully retrieved evidence into the requested
entity-level answer.

\noindent
\textbf{Supporting trace evidence.}
Both preceding tool results succeeded with HTTP status 200, and the
second result reached the Potash Markets website. This rules out an
access failure at the attributed node. The evaluator independently
records \textit{Potash Markets -- Clark Street} as both a reference item
and a missing item. The failure is consequently an incomplete answer,
not an unavailable-resource failure.
\end{masdverdict}

\end{masdgold}


\begin{masdgold}
  {GOLD--C2}
  {Multi-Target and Complex Context}
  {MASDAmber}

\masdmeta
  {1aaa0ae52cbcf325}
  {e5b605fd443d173b}
  {webarena_verified_hard_267}
  {WebArena verified-hard split / AG2}
  {15 nodes; 3 recorded API interactions}
  {webarena_verified_hard_267:attempt_000:snapshot:9ad08821a7a42cde:tool_result_0002}
  {data/initial_failure_logs/ag2/webarena_verified_hard_267/attempt_000}

\begin{masdpanel}[colframe=MASDAmber!65]{Task prompt}
\textit{Get the relation ID of the closest national park to the hometown
of Stephen King and the time to drive there. Return a list of objects
with keys {\upshape\texttt{relation\_id}} (integer) and
{\upshape\texttt{duration}}
(in HH:MM:SS format) only, without any additional details. Use the OSRM
direction service and the provided wiki to look up any needed
information, and search both source and destination by coordinates from
the place official page on the wiki.}
\end{masdpanel}

\begin{masdpanel}{Recorded execution path}
\begin{center}
\resizebox{\linewidth}{!}{%
\begin{tikzpicture}[node distance=3.0mm]
  \node[masdok] (wiki1)
    {Stephen King\\wiki page\\HTTP 200};
  \node[masdloop, right=of wiki1] (coord)
    {coordinate query\\HTTP 200\\no coordinates};
  \node[masdloop, right=of coord] (wiki2)
    {same wiki page\\requested again\\HTTP 200};
  \node[masdfail, right=of wiki2] (final)
    {terminal state\\still a browse\\action};

  \draw[masdarrow] (wiki1) -- (coord);
  \draw[masdarrow] (coord) -- (wiki2);
  \draw[masdarrow] (wiki2) -- (final);

  \draw[
    -{Latex[length=1.7mm]},
    dashed,
    line width=0.6pt,
    draw=MASDAmber
  ]
    ([yshift=1.2mm]wiki2.north)
    to[bend left=32]
    node[above, font=\sffamily\scriptsize, text=MASDAmber]
      {repetition}
    ([yshift=1.2mm]wiki1.north);

  \node[
    below=1.2mm of wiki2,
    text=MASDAmber,
    font=\sffamily\bfseries\scriptsize
  ] {GOLD FAILURE NODE};
\end{tikzpicture}%
}
\end{center}
\end{masdpanel}

\begin{masdpanel}{Field-preserving raw trace excerpt}
\begin{lstlisting}[style=MASDTrace]
case_id = "1aaa0ae52cbcf325"
annotation_id = "e5b605fd443d173b"

failure_node_id =
"webarena_verified_hard_267:attempt_000:
 snapshot:9ad08821a7a42cde:tool_result_0002"

tool_call_0000.url =
"https://en.wikipedia.org/wiki/Stephen_King"

tool_result_0000.status = "success"
tool_result_0000.http_status = 200
tool_result_0000.content_length = 889638

tool_call_0001.query.prop = "coordinates"
tool_result_0001.status = "success"
tool_result_0001.http_status = 200
tool_result_0001.response =
{"batchcomplete":"",
 "query":{"pages":{"26954":
   {"pageid":26954,"ns":0,"title":"Stephen King"}}}}

tool_call_0002.url =
"https://en.wikipedia.org/wiki/Stephen_King"

tool_result_0002.status = "success"
tool_result_0002.http_status = 200
tool_result_0002.content_length = 889638

final_result.action = "browse"
final_result.url =
"https://en.wikipedia.org/wiki/Stephen_King"

evaluator.expected =
[{"duration":"01:33:00","relation_id":2176999}]
\end{lstlisting}
\end{masdpanel}

\begin{masdverdict}{MASDAmber}{Attribution}
\masdpill{MASDAmber}{C2}
\textbf{Repetition, loop, or progress deadlock.}

\noindent
\textbf{Erroneous node.}
The gold locus is \path{tool_result_0002}, which completes a repeated
request for the same Stephen King page after the coordinate query has
already failed to provide coordinates.

\noindent
\textbf{Error committed.}
The task requires a multi-hop chain:
hometown identification, source coordinates, nearest-national-park
identification, destination coordinates, OSRM routing, relation-ID
lookup, and schema-constrained formatting. Instead of advancing to a
new subgoal, the trajectory returns to the previously retrieved page.

\noindent
\textbf{Supporting trace evidence.}
The first and third requests use the same Wikipedia URL and both return
HTTP 200 with the same recorded content length of 889638. The
intermediate coordinate query also returns HTTP 200, but its response
contains page identity fields and no coordinates. The final state
remains a browsing action, whereas the evaluator expects
\texttt{relation\_id=2176999} and
\texttt{duration=01:33:00}. Successful HTTP status therefore does not
explain the failure; the decisive problem is lack of progress.
\end{masdverdict}

\end{masdgold}


\begin{masdgold}
  {GOLD--C3}
  {Incomplete or Low-Signal Operational Context}
  {MASDRed}

\masdmeta
  {06845b3bc9dad059}
  {fadb3bf17df2786e}
  {webarena_verified_hard_320}
  {WebArena verified-hard split / AG2}
  {13 nodes; 3 recorded API interactions}
  {webarena_verified_hard_320:attempt_000:snapshot:9ad08821a7a42cde:tool_result_0000}
  {data/initial_failure_logs/ag2/webarena_verified_hard_320/attempt_000}

\begin{masdpanel}[colframe=MASDRed!60]{Task prompt}
\textit{How much refund should I expect from my orders canceled, if any,
in February 2023, including the shipping fee? Return the value as a
number only, without any additional details.}
\end{masdpanel}

\begin{masdpanel}{Recorded execution path}
\begin{center}
\resizebox{\linewidth}{!}{%
\begin{tikzpicture}[node distance=3.0mm]
  \node[masdstep] (placeholder)
    {request to unresolved\\shopping\\placeholder};
  \node[masdfail, right=of placeholder] (schema)
    {\path{tool_result_0000}\\MissingSchema};
  \node[masdfail, right=of schema] (dns)
    {retry with HTTPS\\name resolution\\failure};
  \node[masdstep, right=of dns] (final)
    {unsupported final\\value\\58.99};

  \draw[masdarrow] (placeholder) -- (schema);
  \draw[masdarrow] (schema) -- (dns);
  \draw[masdarrow] (dns) -- (final);

  \node[
    below=1.2mm of schema,
    text=MASDRed,
    font=\sffamily\bfseries\scriptsize
  ] {GOLD FAILURE NODE};
\end{tikzpicture}%
}
\end{center}
\end{masdpanel}

\begin{masdpanel}{Field-preserving raw trace excerpt}
\begin{lstlisting}[style=MASDTrace]
case_id = "06845b3bc9dad059"
annotation_id = "fadb3bf17df2786e"

failure_node_id =
"webarena_verified_hard_320:attempt_000:
 snapshot:9ad08821a7a42cde:tool_result_0000"

tool_call_0000.url = "__SHOPPING__"
tool_result_0000.status = "failed"
tool_result_0000.error.type = "MissingSchema"
tool_result_0000.error.message =
"Invalid URL '__SHOPPING__': No scheme supplied.
 Perhaps you meant https://__SHOPPING__"

tool_call_0001.url = "https://__SHOPPING__"
tool_result_0001.status = "failed"
tool_result_0001.error.type = "ConnectionError"
tool_result_0001.error.message =
"... Failed to resolve '__shopping__' ..."

final_result = 58.99
evaluator.expected = 406.53
\end{lstlisting}
\end{masdpanel}

\begin{masdverdict}{MASDRed}{Attribution}
\masdpill{MASDRed}{C3}
\textbf{Unresolved execution, access, or environmental context.}

\noindent
\textbf{Erroneous node.}
The earliest decisive failure is
\path{tool_result_0000}, which receives an unresolved environment
placeholder instead of an executable shopping-site URL.

\noindent
\textbf{Error committed.}
The interaction cannot enter the target application. The initial call
fails because the placeholder has no URL scheme; adding
\texttt{https://} does not repair the missing environment binding and
instead produces a name-resolution failure.

\noindent
\textbf{Supporting trace evidence.}
The recorded exceptions explicitly identify
\texttt{MissingSchema} followed by \texttt{ConnectionError}. No order
history is observed before the trajectory emits 58.99, which differs
from the evaluator value of 406.53. The bundle itself is structurally
complete; the low-signal condition lies in the unresolved execution
context captured by the trace, not in missing provenance files.
\end{masdverdict}

\end{masdgold}


\begin{masdgold}
  {GOLD--C4}
  {Plan--Action Inconsistency}
  {MASDPurple}

\masdmeta
  {11772545c348fb2a}
  {f480002d4509ee1e}
  {webarena_verified_hard_204}
  {WebArena verified-hard split / CrewAI}
  {5 nodes; 1 recorded API interaction}
  {webarena_verified_hard_204:attempt_000:snapshot:9ad08821a7a42cde:agent_action_0000}
  {data/initial_failure_logs/crewai/webarena_verified_hard_204/attempt_000}

\begin{masdpanel}[colframe=MASDPurple!65]{Task prompt}
\textit{Get the product name and final price, ordered from low to high, of
the most recent completed order. Return a list of objects with keys
{\upshape\texttt{name}} and {\upshape\texttt{price}} only, without any
additional details.}
\end{masdpanel}

\begin{masdpanel}{Recorded execution path}
\begin{center}
\resizebox{0.91\linewidth}{!}{%
\begin{tikzpicture}[node distance=5.2mm]
  \node[masdstep] (task)
    {Order-history\\lookup task};
  \node[masdambiguous, right=of task] (claim)
    {\path{agent_action_0000}\\claims a final\\answer};
  \node[masdfail, right=of claim] (missing)
    {no real tool event\\no observed\\order record};

  \draw[masdarrow] (task) -- (claim);
  \draw[masdarrow] (claim) -- (missing);

  \node[
    below=1.2mm of claim,
    text=MASDPurple,
    font=\sffamily\bfseries\scriptsize
  ] {GOLD FAILURE NODE};
\end{tikzpicture}%
}
\end{center}
\end{masdpanel}

\begin{masdpanel}{Field-preserving raw trace excerpt}
\begin{lstlisting}[style=MASDTrace]
case_id = "11772545c348fb2a"
annotation_id = "f480002d4509ee1e"

failure_node_id =
"webarena_verified_hard_204:attempt_000:
 snapshot:9ad08821a7a42cde:agent_action_0000"

agent_action_0000 =
Final answer:
[{"name":"Product B","price":9.99},
 {"name":"Product A","price":19.99}]

Evidence needed:
To obtain the actual data, I would browse to the
__SHOPPING_ADMIN__ page...

evaluator.has_real_tool_event = false
evaluator.observed_tool_result = {}

evaluator.expected =
[{"name":"Ida Workout Parachute Pant","price":38.4},
 {"name":"Proteus Fitness Jackshirt","price":45.0}]
\end{lstlisting}
\end{masdpanel}

\begin{masdverdict}{MASDPurple}{Attribution}
\masdpill{MASDPurple}{C4}
\textbf{Plan--action--outcome inconsistency.}

\noindent
\textbf{Erroneous node.}
The failure is attributed to
\path{agent_action_0000}, where the agent simultaneously presents a
purported final answer and admits that it would still need to browse
the shopping administration page to obtain the actual data.

\noindent
\textbf{Error committed.}
The output is superficially plausible: it is valid JSON-like data,
contains the requested keys, and is sorted by price. However, the
action's own explanation contradicts its epistemic status. It claims
specific products and prices before observing any order record.

\noindent
\textbf{Supporting trace evidence.}
The evaluator records
\path{has_real_tool_event=false} and an empty
\path{observed_tool_result}. The fabricated names and prices also differ
from the expected products and prices. Thus, the error is not merely an
incorrect value or formatting defect; it is a mismatch between the
node's claimed answer and the evidence-gathering action that the same
node says remains necessary.
\end{masdverdict}

\end{masdgold}

\section{Supplementary Artifact Contents}
\label{app:artifact-contents}

The code and data supplement includes the complete source code of \sys and the complete \ds dataset used in this study.
